\documentclass[letterpaper,11pt,onecolumn]{IEEEtran}  

\usepackage{mathptmx}   
\usepackage{amsfonts, amssymb}
\usepackage{amsmath}
\usepackage{graphicx}
\usepackage{theorem}
\usepackage{textcomp}
\usepackage{multirow}
\usepackage{tabularx}
\usepackage{array}
\newcolumntype{L}[1]{>{\raggedright\let\newline\\\arraybackslash\hspace{0pt}}m{#1}}
\newcolumntype{C}[1]{>{\centering\let\newline\\\arraybackslash\hspace{0pt}}m{#1}}
\newcolumntype{R}[1]{>{\raggedleft\let\newline\\\arraybackslash\hspace{0pt}}m{#1}}

\makeatletter
\let\NAT@parse\undefined
\makeatletter
\usepackage{natbib}

\let\chapter\section 
\usepackage[linesnumbered,ruled]{algorithm2e}
\SetKwInput{KwIn}{Initial condition}
\SetKwInput{KwResult}{Outcome}

\newtheorem{theorem}{Theorem}

\newcommand{\qed}{\hfill $\Box$\\}

\def\vec#1{\mathbf{#1}}

\def\cop{COP}
\def\copq{QCOP}

\def\op{OP}
\def\top{TOP}

\makeatletter
\newcommand{\customlabel}[2]{%
\protected@write \@auxout {}{\string \newlabel {#1}{{#2}{}}}}
\makeatother

\title{
Correlated Orienteering Problem and it Application to Persistent Monitoring Tasks
}
\author{Jingjin Yu \qquad Mac Schwager \qquad Daniela Rus%
\thanks{J. Yu and D. Rus with the Computer Science and Artificial Intelligence Lab at the Massachusetts Institute of Technology. E-mail: {jingjin,rus}@csail.mit.edu.}%
\thanks{M. Schwager is with the Mechanical Engineering Department at Boston University. E-mail: schwager@bu.edu.}%
\thanks{This work was supported in part by ONR projects N00014-12-1-1000 and N00014-09-1-1051.}%
}

\begin{document}
\maketitle

\begin{abstract} We propose a novel non-linear extension to the Orienteering Problem (\op), called the Correlated Orienteering Problem (\cop). We use \cop\, to model the planning of informative tours (cyclic paths) for the persistent monitoring of a spatiotemporal field with time-invariant spatial correlations, in which the tours are constrained to have fixed-length or -time budgets. Our focus in this paper is \copq, a quadratic \cop\, formulation that only looks at correlations between neighboring nodes in a node network. The main feature of \copq\, is a quadratic utility function that captures the said spatial correlation. \copq\, may be solved using mixed integer quadratic programming (MIQP), with the resulting anytime algorithm capable of planning multiple disjoint tours that maximize the quadratic utility. In particular, our algorithm can quickly plan a near-optimal tour over a network with up to $150$ nodes. Besides performing extensive simulation studies to verify the algorithm's correctness and characterize its performance, we also successfully applied it to two realistic persistent monitoring tasks: {\em (i)} estimation over a synthetic spatiotemporal field, and {\em (ii)} estimating the temperature distribution in the state of Massachusetts. 
\end{abstract}

\section{Introduction} 
\begin{figure}[ht!]
\begin{center}
    \includegraphics[width=3in]{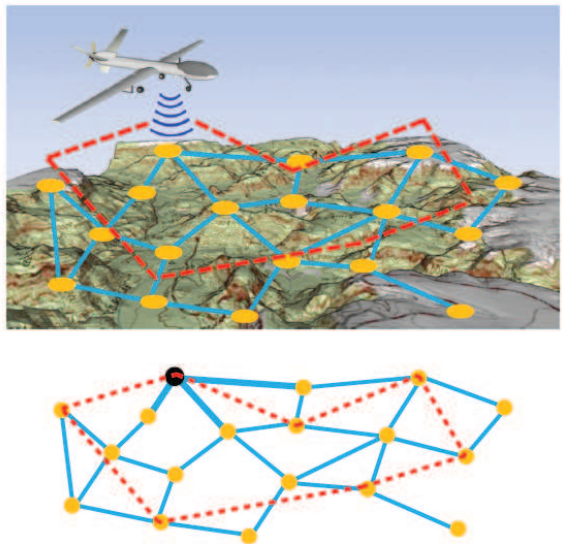}
\end{center}
\caption{\label{fig:example} [top] A surveillance scenario in which an UAV with limited range is faced with the problem of covering a large number of nodes. [bottom] The abstracted node network (dots and solid lines) and the tour (dashed lines) taken by the UAV. As a measurement is made from the UAV at a node, for example the dark node in the figure, through spatial correlation, partial information about its neighbors can also be inferred from the same measurement (in this case, the neighbors are the three nodes connected to the dark node though the three bold edges). Following the dashed tour lines, which is much shorter than a traveling salesman (TSP) tour, the UAV can provide at least partial information about every node in the network.}
\end{figure}

Consider the problem of dispatching unmanned aerial vehicles (UAVs) with on-broad cameras to monitor road traffic in a large city. Often, UAVs have limited range and can stay in air only for a limited amount of time. On the other hand, traffic events such as congestion tend to have strong local correlations, {\em i.e.}, if the vehicle density at an intersection is high, the same is likely true at intersections that are close-by. Therefore, sequentially visiting intersections following the road network's topological structure may offer little incremental information. As UAVs are not restricted to travel along roads, routes with carefully selected, not necessarily adjacent intersections can potentially offer much better overall traffic information per unit of traveled distance. Under such settings, the following question then naturally arises: how to plan the best tours for the UAVs so that they can collect the maximum amount of traffic information per flight? Application scenarios like this are far from unique. For example, nearly identical settings appear when we want to deploy autonomous marine vehicles to collect samples for the detection of water pollution events such as oil spills, or when a political candidate wants to maximize his/her reach given limited travel and time budgets. A graphical example illustrating the settings of such problems is provided in Figure~\ref{fig:example}.

In persistent surveillance and monitoring tasks using mobile robots with onboard sensors, the robots usually have fixed base stations which they must depart from and return to. Moreover, these robots often have limited travel distance or time budget. Thus, when a large number of points of interest (nodes) must be surveyed, it may well be the case that only a subset of the nodes can be visited by the robots. Then, choices among the nodes must be made to accommodate two conflicting goals: {\em (i)} each robot must follow a tour (cyclic path) of which the total cost does not exceed its travel budget, and {\em (ii)} the tours must be planned to maximize the amount of collected information, as measured by some utility (reward) function. When nodes have information utilities that are additive, an {\em Orienteering Problem} (\op) \cite{VanSouVan11}, a problem closely related to the well known {\em Traveling Salesman Problem} (TSP) \cite{Lap92}, is obtained. Research on \op\, has yielded effective algorithms for solving many versions of the problem, including the {\em Team Orienteering Problem} (\top) \cite{ChaGolWas96a}, in which multiple tours must be planned.\footnote{Henceforth, we use {\em Orienteering Problem} (\op) as a blanket term to cover all additive/linear utility orienteering problems, which include \top.} 

In practice, however, the information collected at a node is frequently correlated with the information collected at adjacent nodes, rendering the total utility a non-linear combination of individual node utility. That is, it is often the case that such information can be viewed as forming a locally-correlated spatiotemporal field; surveying a given node will also offer partial information about its neighbors. For example, the nodes may be cities, city blocks, and locations in reservoirs with the associated quantities being population dynamics, criminal activities, and water pollutant concentration, respectively. In this paper, assuming that the spatial correlation among the nodes are intrinsic ({\em i.e.}, determined by local structures and mostly time-invariant) we propose an extension to \op, called the {\em Correlated Orienteering Problem} (\cop), to incorporate such correlations in the informative path planning phase. We do not assume convexity or submodularity over the underlying field. In particular, we focus on \copq, a \cop\, instantiation that only looks at correlations between immediate neighbors. After formulating \cop\, and \copq, we derive mixed integer quadratic programming (MIQP) models for solving the problem for single and multiple robots. Our simulation studies suggest that {\em (i)} \copq\, captures spatial correlations among the nodes quite well, and {\em (ii)} the MIQP-based {\em anytime} algorithm quickly yields approximate solutions to the \copq\, problem for instances of up to $150$ nodes with user-specified optimality bounds. We note that \cop\, and \copq\, are NP-hard problems. 

\textbf{Related work:} Our work fuses ideas from two relatively disjoint branches of research: (discrete) \op\, and (mostly continuous) informative path and policy planning problems. \op, as indicated by its name, has its origin from orienteering games \cite{ChaGolWas96a, ChaGolWas96b}. In such a game, rewards of uniform or varying sizes are spatially scattered. To collect a reward, a player must physically visit the location where the reward is placed to pick it up. The goal for a player or a team of players is to plan the best path(s) to gather the maximum possible reward given limited time. Thus, \op\, can be viewed as a variation of both the Knapsack Problem (KP) \cite{Kar72} and the Traveling Salesman Problem (TSP) \cite{Lap92}. For a detailed account of \op, see \cite{VanSouVan11}. 

The literature on informative path and policy planning for persistent monitoring is fairly rich \cite{MicStuMoh11,SmiSchSmiJonRusSuk11,AlaFatSmi12,ArvKimMar12,CasLinDing13,GirHowHed05,GroKelKumPap06,LanSchwagerICRA13GRFPersistentMonitoring,NigKro05,LeDahFerFra08,SmithSchwagerRusTRO12Persistence,SolteroIROS12PathMorphing,HolSuk14IJRR,YuKarRus14ICRA}, covering theories, systems, algorithm designs, and applications. In the works presented in \cite{AlaFatSmi12,ArvKimMar12,LanSchwagerICRA13GRFPersistentMonitoring,LeDahFerFra08,SmithSchwagerRusTRO12Persistence,HolSuk14IJRR}, fundamental limits as well as provably correct algorithms were established for a variety of persistent monitoring problems. At the same time, comprehensive systems have been designed to address specific application domains, such as ground, aerial, and underwater applications \cite{MicStuMoh11,SmiSchSmiJonRusSuk11,GirHowHed05,GroKelKumPap06,NigKro05}. On work most closely related to ours, in \cite{AlaFatSmi12}, iterative TSP paths are planned to minimize the maximum latency across all nodes in a connected network. The authors show that the approach yields $O(\log n)$ approximation on optimality in which $n$ is the number of nodes in the network. The problem of generating speed profiles for robots along predetermined cyclic paths for keeping bounded the uncertainty of a varying field is addressed in \cite{SmithSchwagerRusTRO12Persistence}, in which the authors characterize appropriate policies for both single and multiple robots. In \cite{SolteroIROS12PathMorphing}, decentralized adaptive controllers were designed to morph the initial closed paths of robots to focus on regions of high importance.

Sampling based planning methods ({\em e.g.} PRM, RRT, RRT$^*$ and their variations \cite{KavSveLatOve96,Lav98c,KarFra11IJRR}) have also been applied to informative path planning problems. In \cite{HolSuk14IJRR}, Rapidly-Exploring Random Graphs (RRG) are combined with {\em branch and bound} methods for planning the most informative path for a mobile robot. In \cite{LanSchwagerICRA13GRFPersistentMonitoring}, the authors tackle the problem of planning a cyclic trajectory for best estimation of a time-varying Gaussian Random Field, using a variation of RRT called Rapidly-Expanding Random Cycles (RRC). 

Lastly, our problem, and \op\, in general, also has a {\em coverage} element. Coverage of a two-dimensional region has been extensively studied in robotics \cite{Cho00,Cho01,GabRim03}, as well as in purely geometric settings, for example, in~ \cite{ChiNta88}, where the proposed algorithms compute the shortest closed routes for continuous coverage of polygonal interiors under an infinite visibility sensing model. Coverage with limited sensing range was also addressed later \cite{HokStiSpo08,Nta91}. 

\textbf{Contribution:} This paper brings three main contributions: 
\begin{itemize}
\item We introduce \cop\, as a novel non-linear extension to \op, to model and harness time-invariant spatial correlations that are frequently present in informative path and policy planning problems. In particular, our formulation addresses the challenging problem of planning information maximizing tours for single and multiple robots under a limited travel budget.
\item We provide complete mixed integer quadratic programming (MIQP) models for solving a \copq, a quadratic instantiation of \cop. These models, combined with a good off-the-shelf MIQP solver, yields an anytime algorithm that can effectively compute robot tour(s) over networks with tens to hundreds of nodes, for the optimal estimation of the underlying spatially corrected spatiotemporal fields. 
\item We demonstrate that our models and algorithms are effective over both simulated and empirical data sets. 
\end{itemize}

In comparison to the conference publication \cite{YuSchwagerRus14IROS}, the current paper provides a much more thorough treatment of \cop. From the perspective of the problem statement, we now give a cleaner and more general formulation. From the perspective of algorithmic solutions, we have developed an anytime algorithm and additional heuristics, which greatly boost the computational speed, allowing much larger problems to be solved. In the simulation study, a much more comprehensive evaluation of the algorithmic performance as well as simulations on real temperature data are now included. 

\textbf{Organization:} The rest of the paper is organized as follows. In Section~\ref{section:formulation}, we formally introduce \cop\, and \copq. In Section~\ref{section:algorithm}, we derive MIQP-based anytime algorithms for solving \copq\, for single and multiple robots. In Sections~\ref{section:experiments}, we perform extensive computational experiments to verify the correctness and evaluate the performance of our algorithmic solutions. We then illustrate how \copq\, may be applied to solve realistic persistent monitoring tasks in Section~\ref{section:application} and conclude the paper in Section~\ref{section:conclusion}. Table~\ref{table_symbols} lists symbols that are frequently used in the paper. 

\begin{table}[htp]
\begin{center}
	\caption{\label{table_symbols} List of frequently used symbols and their interpretations. }
	\begin{tabular}{cl}
	 \hline\hline \\
	 $V = \{v_i\}$ & Node or point of interest, $|V| = n$ \\
	 $G = (V, E)$ & Node network\\
	 $\vec p_i$ & The two dimensional coordinate of $v_i$\\
	 $r_i$ & Utility of knowing complete information about $v_i$\\	 
	 $\psi(v_i, t)$ & Time-varying scalar field\\	 
	 $A = \{a_k\}$ & Mobile robot $k$, $|\{a_k\}| = m$ \\ 
	 $v_{b_k}$ & Start and end node for robot $a_k$ \\
	 $c_k$ & Travel budget for robot $a_k$\\
	 $\Pi$ & $\pi_1, \ldots, \pi_m$, a set of robot tours \\
	 $J(\Pi)$ & The cost function for a given set of robot tours\\
	 $w_{ij}$ & The weight measuring $v_i$'s influence on $v_j$ \\
	 $x_i$ & \parbox[t]{6.5cm}{Binary variable indicating whether $v_i$ is on a tour}\\
	 $x_{ij}$ & \parbox[t]{6.5cm}{Binary variable indicating whether $v_j$ is visited immediately after $v_i$}\\
	 $x_{ijk}$ & \parbox[t]{6.5cm}{Binary variable indicating whether $v_j$ is visited immediately after $v_i$, by robot $a_k$}\\
	 $u_{i}$ & \parbox[t]{6.5cm}{Integer variable, $2 \le u_i \le n$, the order of $v_i$ in a tour path, if used}\\
	 $u_{ik}$ & \parbox[t]{6.5cm}{Integer variable, $2 \le u_{ik} \le n$, the order of $v_i$ in a tour path, if used, by robot $a_k$}\\
	 $d_{ij}$ & Travel cost from $v_i$ to $v_j$, maybe non-symmetric\\
	 $\alpha_{ij}, \beta_{ij}$ & Linear regression coefficients\\
	 \hline\hline
	 \end{tabular}
\end{center}
\end{table}

\section{Problem Statement}\label{section:formulation}
We study the problem of using mobile sensing robots to periodically survey spatially distributed locations (nodes), assuming that the quantities to be measured at the nodes come from a smooth spatiotemporal (scalar or vector) field. Due to spatial and temporal variations, such fields can be highly complex and dynamic. However, in applications involving large spatial domains ({\em e.g.}, terrains, road networks, forests, oceans, and so on), the underlying spatial domain often does not change. The observation allows us to work with the premise that nearby nodes have mostly {\em time-invariant} spatial correlations\footnote{Here, we use the broad meaning of correlation, which could be, but is not necessarily, the correlation of random variables.}, even though the overall field may change significantly over time. Exploiting these correlations, at any given time, it becomes possible to infer the field's value at a certain node from the values of adjacent nodes. 

Before formally stating the problem, roughly speaking, we are interested in deploying mobile sensors with limited travel range to sample nodes of a network. Based on the samples at the nodes, we then infer the field's value at the rest of the nodes from correlation when possible. Besides time-invariant spatial correlations, we assume that the field remains relatively static during a single trip of the mobile sensor(s). We denote this problem the {\em Correlated Orienteering Problem} (\cop). After introducing the broad \cop\, problem, we focus on a special type of \cop\, with a quadratic cost function induced by independent, linear correlations between adjacent nodes. We call this special instantiation of \cop\, the {\em Quadratic Correlated Orienteering Problem} (\copq). Below, \cop\, and \copq\, are formally defined. 

\subsection{Correlated Orienteering Problem}
Let $V = \{v_1, \ldots, v_n\}$ be a set of spatially distributed nodes in some workspace $W \subset \mathbb R^2$. Each node $v_i \in V$ is associated with coordinates $\vec p_i \in \mathbb R^2$. Let $\psi(\vec p, t), \vec p \in \mathbb R^2$ be a time varying scalar field over $W$. The values on the nodes of $V$, with a slight abuse of notation, are written as as $\psi(v_i, t), 1 \le i \le n$. We assume that the spatial relationship among the nodes of $V$, as determined by $\psi$, induces a directed graph $G$ over $V$. More precisely, $G = (V, E)$ has an edge $(v_i, v_j)$ if and only if $\psi(v_j, t)$ is dependent on $\psi(v_i, t)$. That is, let $N_i = \{v_{i_1}, \ldots, v_{i_k}\}$ be the set of neighbors of $v_i$ in $G$, with edges pointing to $v_i$, then for some time-invariant $f_i$, 
\begin{equation}\label{equation:psi}
\psi(v_i, t) = f_i(\psi(v_{i_1}, t), \ldots, \psi(v_{i_k}, t)).
\end{equation}

Let there be $m$ mobile robots, $A = \{a_1, \ldots, a_m\}$. Each robot follows the standard single integrator dynamics with constant magnitude on the control input  ({\em i.e.}, $\dot x = u$ with $\parallel u\parallel = 1$). To model the travel distance constraints inherent with mobile robots, for a robot $a_k$, let its base ({\em i.e.}, where it must start and end in a cyclic tour) be a node $v_{b_k} \in V$. Let
$$
c: A \to \mathbb R^+, a_k \mapsto c_k
$$
represent the maximum distance (budget) the mobile robots can travel before they must return to their respective bases. Other than the single integrator dynamics and the distance budget constraints, these robots have no other motion constraints. In particular, a robot is not constrained to the implicitly defined graph $G$ and can travel in a straight line between any $\vec p_i, \vec p_j \in W$ as permitted by the travel distance budget. 

A mobile robot can measure $\psi(v_i, t)$ when the robot is located at $v_i \in V$. Let $\Pi = \{\pi_1, \ldots, \pi_m\}$ be a set of tours in which each $\pi_k$ is a cyclic path for $a_k$ that goes through a set of nodes of $V$ including $v_{b_k}$. The overall quality of $\Pi$ is measured by some utility function $J: \{\Pi\} \to \mathbb R^+\cup \{0\}$ that maps path sets to non-negative real values. We do not consider sensor measurement noises in the current paper. 

\begin{figure}[ht!]
\begin{center}
    \includegraphics[width=3in]{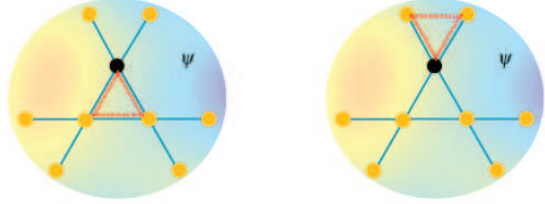}
\end{center}
\caption{\label{fig:cop} Two tours with the same budget over the same node network and $\psi$. Here, all edges have unit length. Given a budget of $3$ with the black node as the starting node, intuitively, with correlations of node values between nearby nodes, if we want to estimate $\psi$ over all $9$ nodes, the tour (dashed path) on the left is likely better because each of the $9$ nodes is adjacent to some node visited by the tour. \cop\, aims to allow the planning of such a tour $\Pi$ through the maximization of a properly defined $J(\Pi)$.}
\end{figure}

The {\em Correlated Orienteering Problem} (\cop) is defined as a 4-tuple $(V, \psi, \{v_{b_k}\}, J)$ over which we wish to find a set of tours $\Pi$ that maximizes the utility $J(\Pi)$. Note that $\psi$ is fixed but generally unknown; it can only be measured (by mobile robots) at particular locations and time instances. An illustrative and qualitative example of what \cop\, aims to achieve is given in Figure~\ref{fig:cop}. 

\textbf{Remark.} At a first glance, \cop\, may appear to mimic a problem whose underlying process is a Markov Decision Process (MDP). Although there are some similarities between the two formulations ({\em e.g.}, like in an MDP-based problem, in \cop, the robots take actions to go to different physical states and the information to be collected follows some distribution), a key difference is that it is never beneficial to revisit a node in a \cop\, instance but revisiting a state in an MDP problem can be rewarding. This is also a key source of computational difficulty associated with \cop\, because dynamic programming techniques that are useful in solving MDP problems can no longer be applied to \cop. 
\subsection{Quadratic Correlated Orienteering Problem }
After defining the general \cop\, problem class, we now describe an instantiation of \cop\, with a quadratic utility (alternatively, reward or cost) function. If the tours of the $m$ robots, $\Pi$, go through a node $v_i \in V$, then a utility of $r(v_i)$ is collected, defined according to the mapping
$$
r: V \to \mathbb R^+, v_i \mapsto r_i.
$$

The robots do not gain more utility for revisiting $v_i$ in $\Pi$. To represent the total utility of \copq, let $\{x_1, \ldots, x_n\}$ be $n$ binary variables, with $x_i = 1$ if and only if $v_i$ ({\em i.e.}, $\vec p_i$) is on some tour $\pi_k \in \Pi$. To incorporate correlations among the nodes during the tour planning phase while also rendering the formulation more concrete, we let $\psi$ (and therefore $\{f_i\}$) and $J$ have the following instantiation. Let the weight function
$$
w: E \to \mathbb R^+, (v_j,v_i) \mapsto w_{ji}
$$
represent the effectiveness of using $\psi(v_j, t)$ to estimate $\psi(v_i, t)$. One may view $w_{ji}$ as representing the amount of information that $\psi(v_j, t)$ has about  $\psi(v_i, t)$, independent of other neighbors of $v_i$. The utility that can be collected over a node $v_i$ is defined as 
\begin{equation}\label{equation:quadratic-objective}
r_i\left(x_i + \sum_{v_j \in N_i} w_{ji}x_j(x_j-x_i)\right), 
\end{equation}
in which the quadratic term $x_j(x_j-x_i)$ is non-zero if and only if $x_j = 1$ and $x_i = 0$. Essentially,~\eqref{equation:quadratic-objective} says that correlations is only relevant for $v_i$ if $v_i$ is not directly visited by a robot. Obviously, for each $0 \le i \le n$, $\sum_{v_j \in N_i} w_{ji} \le 1$. Note that we do not assume that $w_{ij} = w_{ji}$, which may be the case if the field's values at $v_i$ and $v_j$ are, for example, jointly Gaussian random variables.  Over a set of tours for the robots, $\Pi$, the total utility to be maximized is then
\begin{equation}\label{equation:objective}
J(\Pi) = \sum_{i = 1}^n\left(r_ix_i + \sum_{v_j \in N_i} r_jw_{ij}x_i(x_i-x_j)\right). 
\end{equation}

We observe that the utility function~\eqref{equation:objective} defines a natural quadratic extension to \op, which has a linear utility $\sum_{i = 1}^nr_ix_i$. We denote this \cop\, instantiation as the {\em Quadratic Correlated Orienteering Problem} (\copq). By assuming independence among $w_{ij}$'s in formulating \copq, we trade model fidelity for computational efficiency. Nevertheless, we note that \copq\, is still a difficult problem computationally. 

\begin{theorem}\cop\, and \copq\, are NP-hard.\end{theorem}
\noindent {\sc Proof.} It is straightforward to observe that \op, for even a single robot (player), is NP-hard \cite{VanSouVan11}. To see this, for a given travel budget, an algorithm solving \op\, for a single robot must implicitly answer the question of whether the budget is enough for the robot to go through all nodes in $V$. Therefore, \op\, contains as a sub-problem the Euclidean traveling salesman problem (TSP), which is NP-hard. 

For \copq\, with the quadratic utility given by Equation~\eqref{equation:objective}, making the weights $\{w_{ij}\}$ sufficiently small reduces it to an \op, because the quadratic utility (the second summation in Equation~\eqref{equation:objective}) then becomes negligible. This shows that a general \cop\, is also NP-hard.~\qed

\section{Mixed Integer Quadratic Programming Models for Quadratic \cop}\label{section:algorithm}

In this section, we propose a quadratic integer programming model with quadratic utility functions and linear constraints (often known as {\em mixed integer quadratic programming} or MIQP) for solving \copq\, with the utility function given by Equation~\eqref{equation:objective}. We start from the case of a single mobile robot and then move to the case of multiple robots. Then, we discuss the associated algorithmic aspects and provide an algorithm outline. 

\subsection{MIQP Model for a Single Robot}
We start with $m = 1$ (single robot) and adapt the constraints from \cite{VanSouVan11}, which yields a compact model. Whereas our description of the model is self-contained for completeness, more background knowledge on the development of linear \op\, models can be found in \cite{VanSouVan11} and the references within. 

Without loss of generality, let the robot start from $v_1$. Let $x_{ij}$ be a binary variable with $x_{ij} = 1$ if and only if the robot visits $v_j$ immediately after it visits $v_i$. Recall that this does not depend on the existence of an edge between $v_i$ and $v_j$ in $G$. Because it is never beneficial to revisit a node, the robot must only enter and leave any node at most once. This allows us to represent the number of times that the robot enters (resp. leaves) a node $i$ as $\sum_{j = 1, j \ne i}^n x_{ij}$ (resp. $\sum_{j = 1, j \ne i}^n x_{ji}$). Both of these quantities can be at most one. The tour constraint then says that these two quantities must be equal, {\em i.e.}, $\sum_{j = 1, j \ne i}^n x_{ij} = \sum_{j = 1, j \ne i}^n x_{ji}$. Recall that $x_i$ is the binary variable indicating whether $v_i$ is visited, we obtain the following constraints
\begin{equation}\label{equation:sum_rest}
\displaystyle\sum_{j = 1, j \ne i}^n x_{ij} = \sum_{j = 1, j \ne i}^n x_{ji} = x_i\le 1, \quad \forall 2 \le i \le n.
\end{equation}

For $i = 1$, because $v_1$ must be visited, we have
\begin{equation}\label{equation:sum_base}
\displaystyle\sum_{i = 2}^n x_{1i} = \sum_{i = 2}^n x_{i1} = x_1 = 1.
\end{equation}

The constraints~\eqref{equation:sum_rest} and~\eqref{equation:sum_base} ensure that the robot will take a tour starting and ending at $v_1$. They do not, however, prevent multiple disjoint tours from being created. To prevent this from happening, let $2 \le u_i \le n$ be {\em integer} variables for $2 \le i \le n$. If there is a single tour starting from $v_1$, then $u_i$ can be chosen to satisfy the constraints
\begin{equation}\label{equation:no_sub_tour}
\displaystyle u_i - u_j + 1 \le (n - 1)(1 - x_{ij}),\quad 2 \le i, j \le n, i \ne j
\end{equation}

To see that this is true, note that since $u_i - u_j + 1 \le n - 1$ regardless of the values taken by $2 \le u_i, u_j \le n$, \eqref{equation:no_sub_tour} can only be violated if $x_{ij} = 1$. The condition $x_{ij} = 1$ only holds if $v_j$ is visited immediately after $v_i$ is visited. Setting $u_i$ to be the order with which $v_i$ is visited on the tour, if $x_{ij} = 1$, then $u_i - u_j + 1 = 0$, satisfying~\eqref{equation:no_sub_tour}. On the other hand, if there is another tour besides the one starting from $v_1$ and when $v_{ij} = 1$, then the RHS of~\eqref{equation:no_sub_tour} equals zero. For~\eqref{equation:no_sub_tour} to hold, we must have $u_i - u_j + 1 \le 0 \Rightarrow u_i < u_j$. However, this condition cannot hold for all consecutive pairs of nodes on a cycle. Thus,~\eqref{equation:no_sub_tour} enforces that only a single cycle may exist. 

With the introduction of the variables $\{x_{ij}\}$, the tour distance constraint can be enforced via
\begin{equation}\label{equation:tour_distance}
\displaystyle\sum_{i = 1}^n\sum_{j = 1, j \ne i}^n x_{ij}d_{ij} \le c_1,
\end{equation}
in which $d_{ij}$ is the distance from $v_i$ to $v_j$ and $c_1$ is the tour distance constraint for the single (first) robot. Note that the distance $d_{ij}$ needs not to be symmetric. Moreover, it is straightforward to incorporate sensing cost at a node $v_i$ by absorbing that cost into $d_{ij}$ for all $j \ne i$. Alternatively, if the sensing cost is not compatible with the travel cost, an additional cost constraint can be added as well. Putting things together, we obtain a complete MIQP model for \copq, summarized below.  

\begin{equation}\label{equation:miqp}
\begin{array}{l}
\textrm{maximize } J(\Pi) = \displaystyle\sum_{i = 1}^n\left(r_ix_i + \sum_{v_j \in N_i} r_jw_{ij}x_i(x_i-x_j)\right) \\
\textrm{subject to } \\
\begin{array}{ll}
& \displaystyle\sum_{j = 1, j \ne i}^n x_{ij} = \sum_{j = 1, j \ne i}^n x_{ji} = x_i\le 1, \quad \forall 2 \le i \le n\\
& \displaystyle\sum_{i = 2}^n x_{1i} = \sum_{i = 2}^n x_{i1} = x_1 = 1\\
&\displaystyle u_i - u_j + 1 \le (n - 1)(1 - x_{ij}),\quad 2 \le i, j \le n, i \ne j \\
&\displaystyle\sum_{i = 1}^n\sum_{j = 1, j \ne i}^n x_{ij}d_{ij} \le c_1
\end{array}
\end{array}
\end{equation}

\subsection{MIQP Model for Multiple Robots}

Extending a single tour to multiple tours is rather straightforward. To accommodate $m$ robots, the variables $\{x_{ij}\}$ and $\{u_i\}$ are extended to $x_{ijk}$ and $u_{ik}$, with $1 \le k \le m$ representing the robots. Constraints~\eqref{equation:sum_rest} and~\eqref{equation:sum_base} become
\begin{equation}\label{equation:sum_rest_multi}
\displaystyle\sum_{j = 1, j \ne i}^n x_{ijk} = \sum_{j = 1, j \ne i}^n x_{jik} \le 1, \quad \forall 1 \le i \le n, 1 \le k \le m,
\end{equation} 
\begin{equation}\label{equation:sum_rest_multi_2}
\displaystyle\sum_{k = 1}^m\sum_{j = 1, j \ne i}^n x_{ijk} = \sum_{k = 1}^m\sum_{j = 1, j \ne i}^n x_{jik} = x_i\le 1, \quad \forall 1 \le i \le n,
\end{equation}
and
\begin{equation}\label{equation:sum_base_multi}
\displaystyle\sum_{i = 1, i \ne b_k}^n x_{b_kik} = \sum_{i = 1, i \ne b_k}^n x_{ib_kk} = x_{b_k} = 1,\quad 1 \le k \le m.
\end{equation}
Equation~\eqref{equation:sum_rest} splits into Equations~\eqref{equation:sum_rest_multi} and~\eqref{equation:sum_rest_multi_2} because we need Equations~\eqref{equation:sum_rest_multi} to ensure that a node is used by at most one robot. With only~\eqref{equation:sum_rest_multi_2} but not~\eqref{equation:sum_rest_multi}, it can happen that one robot enters a node while a different robot exits the same node, which should not happen. 

The constraints on ${u_{ik}}$ become (for all $1 \le k \le m$)
\begin{equation}\label{equation:no_sub_tour_multi}
\displaystyle u_{ik} - u_{jk} + 1 \le (n - 1)(1 - x_{ijk}), \quad i,j \ne b_k, i \ne j, 1 \le i, j \le n.
\end{equation}

The traveled distance constraint, Equation~\eqref{equation:tour_distance}, becomes
\begin{equation}\label{equation:tour_distance_multi}
\displaystyle\sum_{i = 1}^n\sum_{j = 1, j \ne i}^n x_{ijk}d_{ij} \le c_k,\quad 1 \le k \le m.
\end{equation}

Finally, the utility function Equation~\eqref{equation:objective} remains the same. 

{\bf Remark.} The above MIQP model for \copq\, does not allow two robots to start from the same base. We can easily accommodate such scenarios via modifying Equations~\eqref{equation:sum_rest_multi},~\eqref{equation:sum_rest_multi_2}, and~\eqref{equation:sum_base_multi} accordingly. 

\subsection{Algorithmic Perspectives}
In this subsection, we discuss several important algorithmic aspects that were explored in the implementation of MIQP models, aiming at improving the performance of the algorithm. An algorithm outline is also provided. 
\subsubsection*{Anytime Property} An interesting and extremely useful property, which also comes naturally from our MIQP-based method, is that it leads to an {\em anytime} algorithm. Integer linear and quadratic programming solvers, which are often variations of branch-and-bound algorithms, work with a polytope containing all feasible solutions to the (relaxed) continuous optimization problem. The algorithm functions by braking the polytope into smaller pieces and evaluate the objective function on each of these pieces. As the algorithm progresses, more and more of the initial polytope gets truncated. After some initial steps, a tree structure is built and the leaves of the tree contain active portions of the original feasibility polytope. For each of these pieces (which are again polytopes), in a maximization problem, it is relatively easy to locate a feasible solution with the correct integrality condition ({\em i.e.}, a feasible solution in which binary/integer variables get assigned binary/integer values). The maximum objective value from all these feasible solutions is then a lower bound of the optimal value. At the same time, without respecting the integrality constraints, the maximum achievable objective can also be computed for each leaf polytope. This yields a lower bound on the optimal value. The difference between the two bounds is often referred to as the {\em gap}. An optimal solution is found when this gap reaches zero. If the gap gradually decreases in the execution of a branch-and-bound algorithm, which is the case in our problem, an anytime algorithm is obtained. 

For a difficult optimization problem like \copq, having an anytime algorithm with known optimality gap is beneficial for at least two reasons. First, because it generally takes more and more time for an algorithm to find better and better solutions, having the option to stop early at a sub-optimal solution can save a significant amount of time. This is the {\em anytime} perspective. Second, because we know exactly how optimal our solution is at an arbitrary time instance during the execution process, we may stop running the algorithm and have the confidence that a desired level of optimality is achieved. Note that anytime algorithms are not always equipped with quantitative characterizations of how optimal their current solutions are. Our anytime algorithm, on the other hand, always maintains a good estimation on how optimal the current best solution (also known as the incumbent) is. 

\subsubsection*{Algorithm Outline} Putting things together, we obtain the tour planning and node value estimation algorithm outlined in Algorithm~\ref{algorithm:lae}. In lines~\ref{line:model-setup}-~\ref{line:solve-model}, the MIQP model is set up and solved to obtain the desired set of tours for the robots. The robots then follow the planned tours and collect data as they pass over the nodes on these tours in line~\ref{line:collect-data}. The collected data $\{\psi'(v_i, t_s)\}$ is subsequently updated through correlation to yield the final estimated node values. Note that $\textrm{UpdateNodeEstimate}(\cdot)$ is only determined when \copq\, is connected to a particular spatiotemporal field $\psi(\cdot, \cdot)$ (an example is given in Section~\ref{section:application}). 

\def\algo{{\sc \copq\, Estimation}}
\begin{algorithm}
    \SetKwInOut{Input}{Input}
    \SetKwInOut{Output}{Output}
    \SetKwComment{Comment}{\%}{}
    \Input{$G = (V, E)$: node network, $\vert V \vert = n$ \\ 
		$W = \{w_{ij}\}, 1 \le i, j \le n$: correlation weights \\
		$V_B = \{v_{b_1}, \ldots, v_{b_m}\}$: base nodes \\ 
		$C = \{c_1, \ldots, c_m\}$: travel distance budgets \\
		$gap$: optimality tolerance \\ }
    \Output{$\psi'(v_i, t_s), 1 \le i \le n$: node estimation at time $t_s$}

\vspace{0.1in}
		
\Comment{\small Compute robot tours}

\vspace{0.025in}

$M \leftarrow \textrm{SetUpModel}(G, W, V_B, C)$ \label{line:model-setup}\Comment*{\small Set up model}

$\{\pi_1,\ldots,\pi_{b_m}\} \leftarrow \textrm{SolveModel}(M, gap)$\label{line:solve-model}\Comment*{\small Compute tours}

\vspace{0.1in}

\Comment{\small Run tours and collect data}
$\{\psi'(v_i, t_s)\} \leftarrow \textrm{CollectData}(\{\pi_1,\ldots,\pi_{b_m}\})$\label{line:collect-data}

\vspace{0.1in}

\Comment{\small Estimate value at unvisited nodes}
$\{\psi'(v_i, t_s)\} \leftarrow \textrm{UpdateNodeEstimate}(\{\psi'(v_i, t_s)\})$\label{line:estimate}

\vspace{0.1in}

\Return{$\{\psi'(v_i, t_s)\}$}
\caption{\algo} \label{algorithm:lae}
\end{algorithm}

\subsubsection*{Encoding Utility in Linear Form} We observe that the utility given by~\eqref{equation:objective} can in fact be turned into a linear utility without loss of accuracy. To do so, for each $(i, j)$ pair, define an additional binary variable $z_{ij}$. We may then enforce $z_{ij} = 1$ if and only if $x_i(x_i - x_j) = 1$ by adding two constraints,
\begin{align}
z_{ij} &\le x_i, \\
z_{ij} &\le \frac{x_i - x_j + 1}{2},
\end{align}
to our model. Clearly, if $x_i = 0$ or $x_j = 1$, these constraints ensure that $z_{ij} = 0$. On the other hand, updating~\eqref{equation:objective} to 
\begin{equation}\label{equation:objective-linear}
J(\Pi) = \sum_{i = 1}^n\left(r_ix_i + \sum_{v_j \in N_i} r_jw_{ij}z_{ij}\right),
\end{equation}
when $x_i = 1$ and $x_j = 0$, maximizing $J(\Pi)$ ensures that $z_{ij} = 1$. 

Having a linear utility effectively transforms our MIQP model into a mixed integer linear programming (MILP) model. Whereas the transformation does not directly reduce problem complexity, it can be beneficial for the solver to be aware that the problem is in fact an MILP. 

\subsubsection*{Restricting the Travel Distance Between Nodes} During the computational evaluation of our algorithm, we observe that, when utilities of the nodes are relatively similar, an optimal robot path rarely moves between two nodes that are too far away. In fact, due to the local correlation model, when an optimal path moves from one node to another, the second node is almost always within the 2-neighborhood\footnote{A node $v_j$ is in the {\em 2-neighborhood} of a node $v_i$ if there exists $v_k \in N_i$ ($N_i$ is the set of immediate neighbors of $v_i$) such that $v_j \in N_j$.} of the first node.  Based on this observation, we attempted a heuristic that restricts how far a robot may travel from one node to the next. With the heuristic, each node then creates a constant number $x_{ij}$ variables in the MIQP model. Thus, the size of the MIQP model is greatly reduced with the introduction of this heuristic.

\section{Model Correctness and Performance}\label{section:experiments}
In this section, through computational experiments, we further validate the correctness of our MIQP models for \copq\, and evaluate  computational performance of these models. Note that in this section our focus is on $J(\Pi)$ and we do not actually collect data and perform node value estimation ({\em i.e.}, effectively, we only run lines~\ref{line:model-setup}-~\ref{line:solve-model} of Algorithm~\ref{algorithm:lae}). Unless otherwise stated, the quadratic utility~\eqref{equation:objective} is used instead of the linear utility~\eqref{equation:objective-linear}. We implemented the linear and quadratic models in the JAVA programming language interfacing with the Gurobi solver \cite{gurobi}. All computations were performed on a Dual-Intel Xeon E5-2623 workstation under an 8GB JavaVM. 

\subsection{Model Correctness}\label{subsection:simulation-correctness}
\subsubsection*{Regular Grids} 
\begin{figure}[htp]
\begin{center}
    \includegraphics[width=3.35in]{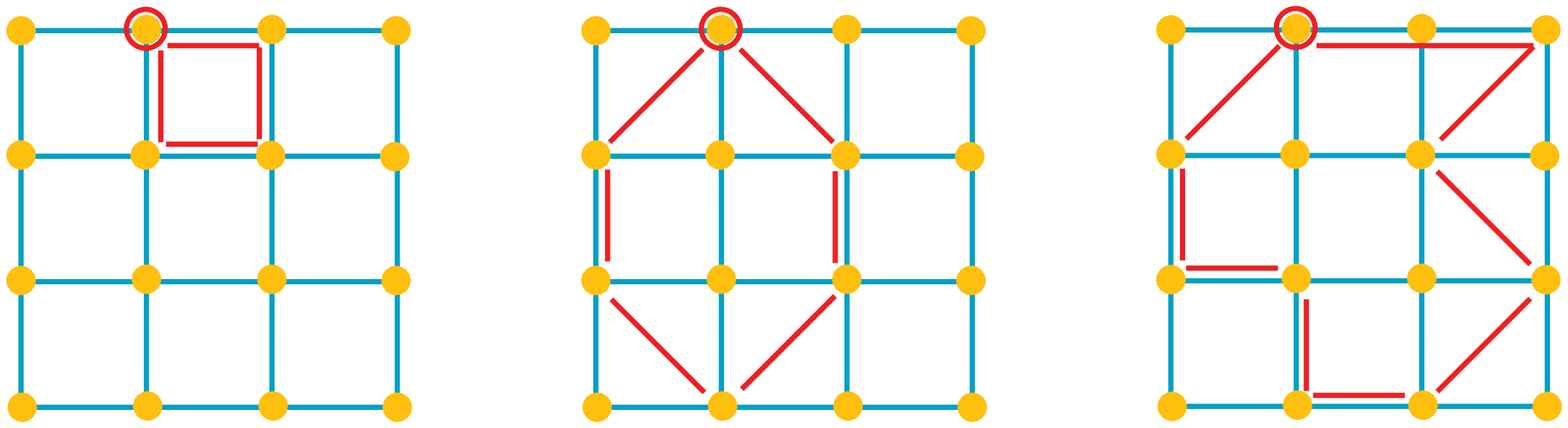}
\end{center}
\caption{\label{fig:4x4-single} Single robot tours for travel budgets $4, 8$, and $12$, in that order, from left to right.}
\end{figure}

We first briefly show that our MIQP model-based algorithms indeed maximize the objective function given by~\eqref{equation:objective}. We begin with a single robot and use $3\times 3$ and $4 \times 4$ grid networks with unit edge length as the test node networks. We set the weights $w_{ji}$ for a node $i$ simply as $1/|N_i|$. For example, if node $i$ has three neighbors, then all $w_{ji}$'s are set to $1/3$. 
\begin{figure}[htp]
\begin{center}
    \includegraphics[width=3.35in]{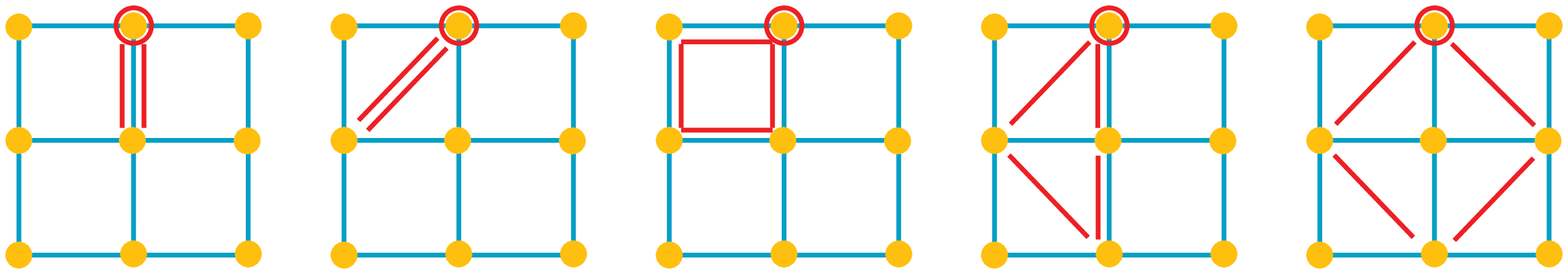}
\end{center}
\caption{\label{fig:3x3-single} Single robot tours for travel budgets $2, 3, 4, 5$, and $6$, in that order, from left to right.}
\end{figure}
For the $3 \times 3$ grid, we let the single robot start at the middle node on the top row (the circled node in Figure \ref{fig:3x3-single}) and let the maximum allowed travel budget vary from $2$ to $6$ with unit increments. Each node has a unit utility. The computed tours for these budgets are illustrated in Figure~\ref{fig:3x3-single}, with utilities as $4.0, 4.5, 5.7, 7.3$, and $9$ (maximum possible), respectively. One can easily verify that these are consistent with the design of the single-robot MIQP model. For the $4\times 4$ grid, under a similar setup, we get the tours as illustrated in Figure~\ref{fig:4x4-single} for travel budgets $4, 8$, and $12$, respectively. The associated maximum utilities are $6.2, 11.5$, and $16.0$, respectively. 

\begin{figure}[htp]
\begin{center}
    \includegraphics[width=3.35in]{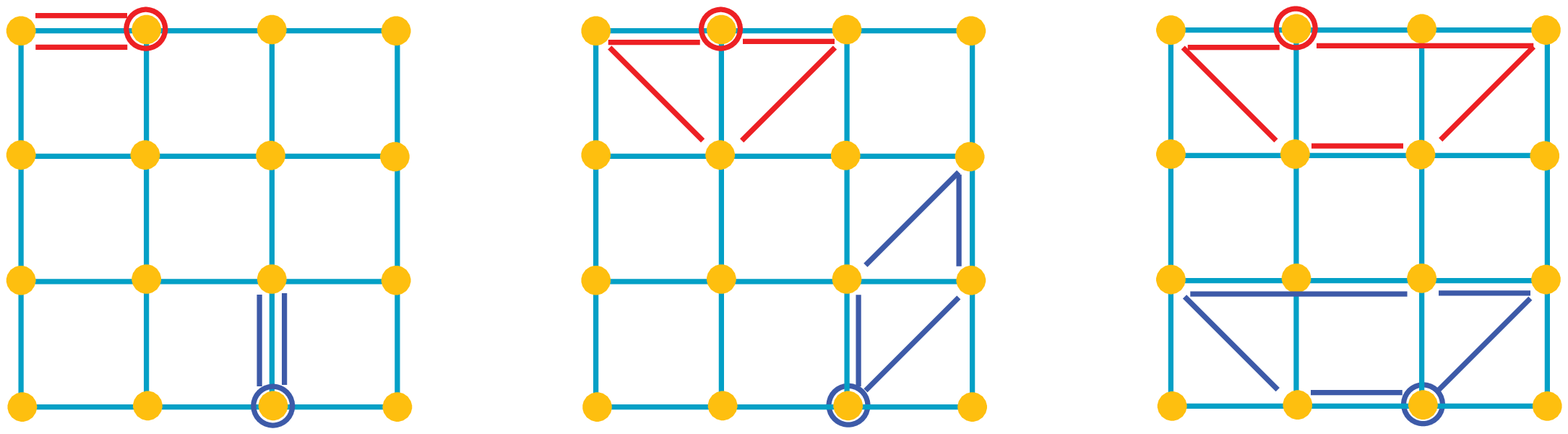}
\end{center}
\caption{\label{fig:4x4-two} Two-robot tours for travel budgets (per robot) $3, 5$, and $7$, in that order, from left to right.}
\end{figure}

Next, a two-robot setup is tested on the $4\times 4$ grid, with the robots starting at opposite locations as indicated by the red and purple circled nodes in Figure~\ref{fig:4x4-two}, which illustrates the tours with individual travel budgets $3, 5$, and $7$, respectively. The associated maximum utilities are $7.2, 12.5$, and $16.0$, respectively. Each problem instance in this subsection took at most two seconds to solve.  

\subsubsection*{Irregular Node Network} 
\begin{figure}[htp]
\begin{center}
    \includegraphics[width=3.5in]{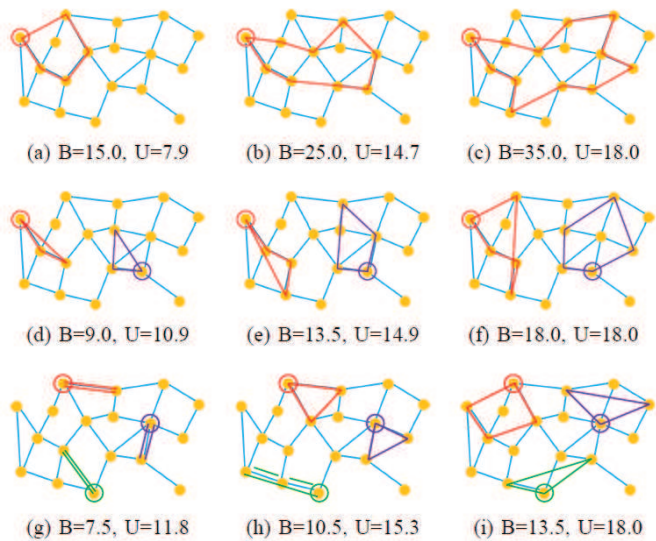}
\end{center}
\caption{\label{fig:irregular} Results from running MIQP models for \copq\, on the irregular, realistic node network from Figure~\ref{fig:example}. The numbers under each picture indicate the budget (B) per tour/robot and the total utility (U), respectively.}
\end{figure}

Our second experiment works with the irregular node network from Figure~\ref{fig:example}. The bounding rectangle of the network is roughly $13$ units by $8$ units. For this network, weights ($w_{ij}$'s) are again computed based on the number of neighbors. Up to three robots were attempted with the longest running time being about $100$ seconds. The trial results and the associated parameters are given in Figure~\ref{fig:irregular}. The base nodes, indicated as colored circles, were hand picked (only once, {\em i.e.}, we did not try any other choices and then select the best one) to be roughly evenly distributed on the network. 

\begin{figure}[htp]
\begin{center}
    \includegraphics[width=1.2in]{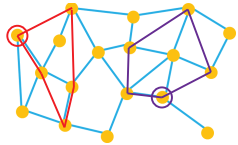}
\end{center}
\caption{\label{fig:irregular-2} Updated two-robot tours with the same utility as the tours in Figure~\ref{fig:irregular}(f).}
\end{figure}

From the result (Figure~\ref{fig:irregular}), we see that the MIQP model always selects tours that do not have spatial overlaps, which is expected but a nice feature to have nevertheless. Also, regardless of the number of robots and tours, the total travel budgets ($35.0$ for one tour, $36.0$ for two tours, and $40.5$ for three) to ensure full coverage of the network appear to be similar. We note that for the cases with multiple tours, some of the individual budgets can be shortened. For example, the tours in Figure~\ref{fig:irregular}(f) can be (manually) updated to the tours in Figure~\ref{fig:irregular-2}, with reduced total (actual) travel budget but without reducing the collected utility. Varying individual budgets is supported by our model by default. 

\subsection{Computational Performance, Regular Grid}
\subsubsection*{Computing Exact (Optimal) Solutions} We now look at the performance of our MIQP/MIP models for solving \copq, starting with computing the exact solution. For all computations, we set a $2,500$-second time limit. To evaluate the computational performance, we again start with a single robot and attempt grid networks with different sizes. For networks with up to $6 \times 6$ nodes, our MIQP model can compute optimal solutions for all choices of budgets within $10$ minutes. We note that a $36$-node network is applicable to many realistic scenarios as demonstrated in Section~\ref{section:application}. The performance characteristics are listed in Table~\ref{table:single_regular}. 

\begin{table}[htp]
\begin{center}
	 \caption{\label{table:single_regular}Performance, regular grids and a single robot}
	 	\begin{tabularx}{\columnwidth}{C{10mm}L{20mm}C{12mm}C{12mm}C{12mm}C{12mm}}
   \hline\hline
	 \multirow{2}*{Grid} & & \multicolumn{4}{c}{Trial $\#$} \\
	 \cline{3-6}
& &  1 & 2 & 3 & 4 \\ \hline
$4\times4$ & \begin{tabular}{c}budget/utility\\time(s)\end{tabular}	
    &	\begin{tabular}{c}3.2/4.3\\0.06\end{tabular}
    &	\begin{tabular}{c}6.4/9.7\\0.40\end{tabular}
		&	\begin{tabular}{c}9.6/13.7\\0.72\end{tabular}
		&	\begin{tabular}{c}12.8/16.0\\0.03\end{tabular} \\ \hline
$5\times5$ & \begin{tabular}{c}budget/utility\\time(s)\end{tabular}	
    &	\begin{tabular}{c}8.0/12.1\\2.4\end{tabular}
    &	\begin{tabular}{c}12.0/18.0\\8.9\end{tabular}
		&	\begin{tabular}{c}16.0/23.3\\5.1\end{tabular}
		&	\begin{tabular}{c}20.0/25.0\\0.3\end{tabular} \\ \hline
$6\times6$ & \begin{tabular}{c}budget/utility\\time(s)\end{tabular}	
    &	\begin{tabular}{c}9.6/14.2\\63.6\end{tabular}
    &	\begin{tabular}{c}14.4/22.3\\45.3\end{tabular}
		&	\begin{tabular}{c}19.2/28.3\\243\end{tabular}
		&	\begin{tabular}{c}24.0/34.0\\217\end{tabular} \\ \hline
$7\times7$ & \begin{tabular}{c}budget/utility\\time(s)\end{tabular}	
    &	\begin{tabular}{c}11.2/16.5\\550\end{tabular}
    &	\begin{tabular}{c}16.8/-\\ $>$2,500\end{tabular}
		&	\begin{tabular}{c} \end{tabular}
		&	\begin{tabular}{c} \end{tabular} \\
	 \hline\hline
	 \end{tabularx}
\end{center}
\end{table}

Because the MIQP model requires one extra set of variables for each extra robot, computing exact solutions for multiple robots becomes more challenging as the number of robots increases. Table~\ref{table:two_regular} lists the complete computational results on the same grids for two robots and up to $5 \times 5$ grids with various budgets. It becomes fairly infeasible to compute for grids of size $6 \times 6$ and beyond with two robots. For three robots (we omit the limited detail here), it becomes challenging to compute exact solutions for all budgets over a $5 \times 5$ grid. 

\begin{table}[htp]
\begin{center}
	 \caption{\label{table:two_regular}Performance, regular grids and two robots}
	 	\begin{tabularx}{\columnwidth}{C{10mm}L{20mm}C{12mm}C{12mm}C{12mm}C{12mm}}
   \hline\hline
	 \multirow{2}*{Grid} & & \multicolumn{4}{c}{Trial $\#$} \\
	 \cline{3-6}
& &  1 & 2 & 3 & 4 \\ \hline
$4\times4$ & \begin{tabular}{c}budget/utility\\time(s)\end{tabular}	
    &	\begin{tabular}{c}3.2/7.2\\0.08\end{tabular}
    &	\begin{tabular}{c}4.8/10.9\\0.17\end{tabular}
		&	\begin{tabular}{c}6.4/14.6\\5.3\end{tabular}
		&	\begin{tabular}{c}8.0/16.0\\0.06\end{tabular} \\ \hline
$5\times5$ & \begin{tabular}{c}budget/utility\\time(s)\end{tabular}	
    &	\begin{tabular}{c}4.0/11.1\\0.11\end{tabular}
    &	\begin{tabular}{c}6.0/15.8\\34.8\end{tabular}
		&	\begin{tabular}{c}8.0/20.5\\1070\end{tabular}
		&	\begin{tabular}{c}10.0/24.3\\1961\end{tabular} \\ 
	 \hline\hline
	 \end{tabularx}
\end{center}
\end{table}

\begin{figure}[htp]
\begin{center}
    \includegraphics[width=1.05in]{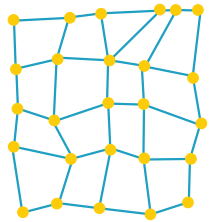}
\end{center}
\caption{\label{fig:5x5-ran} A perturbed $5 \times 5$ grid network.}
\end{figure}

To get a better grasp at the overall performance of the exact algorithm, we randomly perturb regular grids to obtain node networks such as the one shown in Figure~\ref{fig:5x5-ran}. This allows us to get random test cases that mimic more realistic scenarios. Since a regular $7 \times 7$ grid proves to be time-consuming to work with, we focus on network with up to $36$ nodes. The result is listed in Table~\ref{table:single_random}. For each grid size and budget combination, the utility and the computation time are averages over ten randomly generated instances. Comparing Table~\ref{table:single_regular} with Table~\ref{table:single_random}, we observe that instances with randomized grids and regular grids seem to have similar computational complexity. 

\begin{table}[htp]
\begin{center}
	 \caption{\label{table:single_random}Performance, random grids and a single robot}
	 	\begin{tabularx}{\columnwidth}{C{10mm}L{20mm}C{12mm}C{12mm}C{12mm}C{12mm}}
   \hline\hline
	 \multirow{2}*{Grid} & & \multicolumn{4}{c}{Trial $\#$} \\
	 \cline{3-6}
& &  1 & 2 & 3 & 4 \\ \hline
$4\times4$ & \begin{tabular}{c}budget/utility\\time(s)\end{tabular}	
    &	\begin{tabular}{c}3.6/5.1\\0.05\end{tabular}
    &	\begin{tabular}{c}7.1/10.7\\0.90\end{tabular}
		&	\begin{tabular}{c}10.7/15.4\\0.41\end{tabular}
		&	\begin{tabular}{c}14.2/16\\0.14\end{tabular} \\ \hline
$5\times5$ & \begin{tabular}{c}budget/utility\\time(s)\end{tabular}	
    &	\begin{tabular}{c}4.4/6.3\\0.19\end{tabular}
    &	\begin{tabular}{c}8.9/13.5\\9.9\end{tabular}
		&	\begin{tabular}{c}13.3/20.3\\16.0\end{tabular}
		&	\begin{tabular}{c}17.8/25.0\\3.6\end{tabular} \\ \hline
$6\times6$ & \begin{tabular}{c}budget/utility\\time(s)\end{tabular}	
    &	\begin{tabular}{c}10.7/16.4\\195\end{tabular}
    &	\begin{tabular}{c}16/25.6\\177\end{tabular}
		&	\begin{tabular}{c}21.3/32.7\\83.7\end{tabular}
		&	\begin{tabular}{c}26.7/36\\3.7\end{tabular} \\ 
	 \hline\hline
	 \end{tabularx}
\end{center}
\end{table}

\subsubsection*{Performance of Anytime Algorithms} Once we are willing to allow a small amount of deviation from the true optimal solution, which is perfectly acceptable for practical purposes, the anytime property can greatly boost the computational performance. 
In our tests, we set the optimality gap to be $0.2$, meaning that the solution is at most $20\%$ worse than the optimal solution (note that we may still stop the algorithm at any time and have an intermediate solution). Our first experiment compares the anytime algorithm with the exact algorithm by executing the program over the same set of instances used for obtaining the result listed in Table~\ref{table:single_random}. The performance of the anytime algorithm is listed in Table~\ref{table:single_random_anytime}. In comparison to Table~\ref{table:single_random}, the loss of optimality is often much less than the set $20\%$ threshold. At the same time, the computation time is improved up to $30$ times. The improvement become more obvious as the grid size grows. 

\begin{table}[htp]
\begin{center}
	 \caption{\label{table:single_random_anytime}Performance, anytime, random grids and a single robot}
	 	\begin{tabularx}{\columnwidth}{C{10mm}L{20mm}C{12mm}C{12mm}C{12mm}C{12mm}}
   \hline\hline
	 \multirow{2}*{Grid} & & \multicolumn{4}{c}{Trial $\#$} \\
	 \cline{3-6}
& &  1 & 2 & 3 & 4 \\ \hline
$4\times4$ & \begin{tabular}{c}budget/utility\\time(s)\end{tabular}	
    &	\begin{tabular}{c}3.6/5.1\\0.01\end{tabular}
    &	\begin{tabular}{c}7.1/10.6\\0.52\end{tabular}
		&	\begin{tabular}{c}10.7/14.1\\0.22\end{tabular}
		&	\begin{tabular}{c}14.2/14.9\\0.07\end{tabular} \\ \hline
$5\times5$ & \begin{tabular}{c}budget/utility\\time(s)\end{tabular}	
    &	\begin{tabular}{c}4.4/6.2\\0.15\end{tabular}
    &	\begin{tabular}{c}8.9/13.2\\5.2\end{tabular}
		&	\begin{tabular}{c}13.3/19.2\\3.4\end{tabular}
		&	\begin{tabular}{c}17.8/21.7\\0.7\end{tabular} \\ \hline
$6\times6$ & \begin{tabular}{c}budget/utility\\time(s)\end{tabular}	
    &	\begin{tabular}{c}10.7/16.0\\43.7\end{tabular}
    &	\begin{tabular}{c}16/24.3\\13.3\end{tabular}
		&	\begin{tabular}{c}21.3/31.6\\2.9\end{tabular}
		&	\begin{tabular}{c}26.7/31.2\\1.7\end{tabular} \\ 
	 \hline\hline
	 \end{tabularx}
\end{center}
\end{table}

The speed improvement only widens as the problems get bigger. In Table~\ref{table:single-regular-anytime}, we test the anytime algorithm on regular-grid instances (the rest of our tests in this section work only with regular grids, since they appear to be of similar difficulty as instances over random grids) with up to about $150$ nodes. All instances were solved with the longest running time being about $12$ minutes. The result suggests that if we settle for near-optimal solutions, very large problem instances can be solved quickly. 

\begin{table}[htp]
\begin{center}
	 \caption{\label{table:single-regular-anytime}Performance, anytime, regular grids and a single robot}
	 	\begin{tabularx}{\columnwidth}{C{10mm}L{20mm}C{12mm}C{12mm}C{12mm}C{12mm}}
   \hline\hline
	 \multirow{2}*{Grid} & & \multicolumn{4}{c}{Trial $\#$} \\
	 \cline{3-6}
& &  1 & 2 & 3 & 4 \\ \hline
$6\times6$ & \begin{tabular}{c}budget/utility\\time(s)\end{tabular}	
    &	\begin{tabular}{c}10.7/15.5\\16.3\end{tabular}
    &	\begin{tabular}{c}16.0/22.9\\5.0\end{tabular}
		&	\begin{tabular}{c}21.3/30.2\\3.8\end{tabular}
		&	\begin{tabular}{c}26.7/32.5\\1.5\end{tabular} \\ \hline
$8\times8$ & \begin{tabular}{c}budget/utility\\time(s)\end{tabular}	
    &	\begin{tabular}{c}14.2/19.4\\112\end{tabular}
    &	\begin{tabular}{c}21.3/31.3\\169\end{tabular}
		&	\begin{tabular}{c}28.4/39.7\\162\end{tabular}
		&	\begin{tabular}{c}35.6/50.6\\17.3\end{tabular} \\ \hline
$10\times10$ & \begin{tabular}{c}budget/utility\\time(s)\end{tabular}	
    &	\begin{tabular}{c}17.8/26.3\\217\end{tabular}
    &	\begin{tabular}{c}26.7/39.6\\546\end{tabular}
		&	\begin{tabular}{c}35.6/49.3\\236\end{tabular}
		&	\begin{tabular}{c}44.4/64.2\\270\end{tabular} \\ \hline
$12\times12$ & \begin{tabular}{c}budget/utility\\time(s)\end{tabular}	
    &	\begin{tabular}{c}21.3/31.7\\722\end{tabular}
    &	\begin{tabular}{c}32.0/48.5\\235\end{tabular}
		&	\begin{tabular}{c}42.7/61.7\\669\end{tabular}
		&	\begin{tabular}{c}53.3/79.5\\300\end{tabular} \\ 
	 \hline\hline
	 \end{tabularx}
\end{center}
\end{table}

With the anytime algorithm, more challenging multi-robot problems can be solved. For two robots, with $20\%$ optimality tolerance, regular grids with sizes up to $7 \times 7$ can now be tackled. The performance is similar for three robots. The results are listed in Tables~\ref{table:two-regular-anytime} and~\ref{table:three-regular-anytime}.

\begin{table}[htp]
\begin{center}
	 \caption{\label{table:two-regular-anytime}Performance, anytime, regular grids and two robots}
	 	\begin{tabularx}{\columnwidth}{C{10mm}L{20mm}C{12mm}C{12mm}C{12mm}C{12mm}}
   \hline\hline
	 \multirow{2}*{Grid} & & \multicolumn{4}{c}{Trial $\#$} \\
	 \cline{3-6}
& &  1 & 2 & 3 & 4 \\ \hline
$5\times5$ & \begin{tabular}{c}budget/utility\\time(s)\end{tabular}	
    &	\begin{tabular}{c}4.0/10.3\\0.06\end{tabular}
    &	\begin{tabular}{c}6.0/15.8\\13.2\end{tabular}
		&	\begin{tabular}{c}8.0/19.8\\9.6\end{tabular}
		&	\begin{tabular}{c}10.0/21.1\\3.0\end{tabular} \\ \hline
$6\times6$ & \begin{tabular}{c}budget/utility\\time(s)\end{tabular}	
    &	\begin{tabular}{c}4.8/10.9\\0.16\end{tabular}
    &	\begin{tabular}{c}7.2/19.3\\225\end{tabular}
		&	\begin{tabular}{c}9.6/25.3\\322\end{tabular}
		&	\begin{tabular}{c}12.0/30.0\\18.1\end{tabular} \\ \hline
$7\times7$ & \begin{tabular}{c}budget/utility\\time(s)\end{tabular}	
    &	\begin{tabular}{c}5.6/14.6\\144\end{tabular}
    &	\begin{tabular}{c}8.4/23.0\\1329\end{tabular}
		&	\begin{tabular}{c}11.2/30.9\\1640\end{tabular}
		&	\begin{tabular}{c}14.0/37.6\\871\end{tabular} \\ 
	 \hline\hline
	 \end{tabularx}
\end{center}
\end{table}

\begin{table}[htp]
\begin{center}
	 \caption{\label{table:three-regular-anytime}Performance, anytime, regular grids and three robots}
	 	\begin{tabularx}{\columnwidth}{C{10mm}L{20mm}C{12mm}C{12mm}C{12mm}C{12mm}}
   \hline\hline
	 \multirow{2}*{Grid} & & \multicolumn{4}{c}{Trial $\#$} \\
	 \cline{3-6}
& &  1 & 2 & 3 & 4 \\ \hline
$5\times5$ & \begin{tabular}{c}budget/utility\\time(s)\end{tabular}	
    &	\begin{tabular}{c}2.7/10.3\\0.05\end{tabular}
    &	\begin{tabular}{c}4.0/15.3\\0.11\end{tabular}
		&	\begin{tabular}{c}5.3/19.3\\1.1\end{tabular}
		&	\begin{tabular}{c}6.7/21.7\\0.89\end{tabular} \\ \hline
$6\times6$ & \begin{tabular}{c}budget/utility\\time(s)\end{tabular}	
    &	\begin{tabular}{c}3.2/11.3\\0.08\end{tabular}
    &	\begin{tabular}{c}4.8/17.0\\0.44\end{tabular}
		&	\begin{tabular}{c}6.4/25.8\\82.8\end{tabular}
		&	\begin{tabular}{c}8.0/30.3\\26.5\end{tabular} \\ \hline
$7\times7$ & \begin{tabular}{c}budget/utility\\time(s)\end{tabular}	
    &	\begin{tabular}{c}3.7/14.2\\0.2\end{tabular}
    &	\begin{tabular}{c}5.6/22.4\\514\end{tabular}
		&	\begin{tabular}{c}7.5/31.8\\514\end{tabular}
		&	\begin{tabular}{c}9.4/-\\$>$2,500\end{tabular} \\ 
	 \hline\hline
	 \end{tabularx}
\end{center}
\end{table}

\subsubsection*{Linear Utility Function and Restricted Travel Distance} 
With everything being equal, having a linear model instead of a quadratic one generally boosts computational performance because solvers for linear models are more efficient in general. On the other hand, in our particular case, turning the MIQP model to a MILP model introduces a significant number (usually about one-third more) of additional variables, which unfortunately increases the size of the model quite a lot. In practice, we noticed that linear utility by itself does not improve the computational performance by much, unless the problem instance is large ({\em i.e.} $> 50$ nodes). Even though a several-fold speed boost was observed (we omit the limited result here), these instances still take hours to compute using a MILP formulation. 

We also experimented heavily on restricting the travel distance between nodes. Somewhat surprisingly, the reduction of model size from this heuristic, usually over several folds, does not substantially alter the running time. Interestingly, however, when we combine these two heuristics with the anytime algorithm, something of practical importance came out. For instances with up to $100$ nodes, significant performance increase is observed. The result is listed in Table~\ref{table:single-regular-anytime-linear-res}. When compared with Table~\ref{table:single-regular-anytime}, we observe that the computation time is greatly reduced without noticeable impact on the solution quality. 

\begin{table}[htp]
\begin{center}
	 \caption{\label{table:single-regular-anytime-linear-res}Performance, anytime with heuristics, regular grids and a single robot}
	 	\begin{tabularx}{\columnwidth}{C{10mm}L{20mm}C{12mm}C{12mm}C{12mm}C{12mm}}
   \hline\hline
	 \multirow{2}*{Grid} & & \multicolumn{4}{c}{Trial $\#$} \\
	 \cline{3-6}
& &  1 & 2 & 3 & 4 \\ \hline
$6\times6$ & \begin{tabular}{c}budget/utility\\time(s)\end{tabular}	
    &	\begin{tabular}{c}10.7/15.3\\7.5\end{tabular}
    &	\begin{tabular}{c}16.0/23.8\\1.8\end{tabular}
		&	\begin{tabular}{c}21.3/30.0\\0.9\end{tabular}
		&	\begin{tabular}{c}26.7/30.8\\0.1\end{tabular} \\ \hline
$8\times8$ & \begin{tabular}{c}budget/utility\\time(s)\end{tabular}	
    &	\begin{tabular}{c}14.2/20.7\\13.8\end{tabular}
    &	\begin{tabular}{c}21.3/31.2\\10.2\end{tabular}
		&	\begin{tabular}{c}28.4/41.3\\18.2\end{tabular}
		&	\begin{tabular}{c}35.6/51.5\\11.1\end{tabular} \\ \hline
$10\times10$ & \begin{tabular}{c}budget/utility\\time(s)\end{tabular}	
    &	\begin{tabular}{c}17.8/26.3\\24.8\end{tabular}
    &	\begin{tabular}{c}26.7/39.7\\13.1\end{tabular}
		&	\begin{tabular}{c}35.6/52.4\\89.6\end{tabular}
		&	\begin{tabular}{c}44.4/64.1\\167\end{tabular} \\ 
	 \hline\hline
	 \end{tabularx}
\end{center}
\end{table}

\begin{figure*}[tb!]
\begin{center}
    \includegraphics[width=7in]{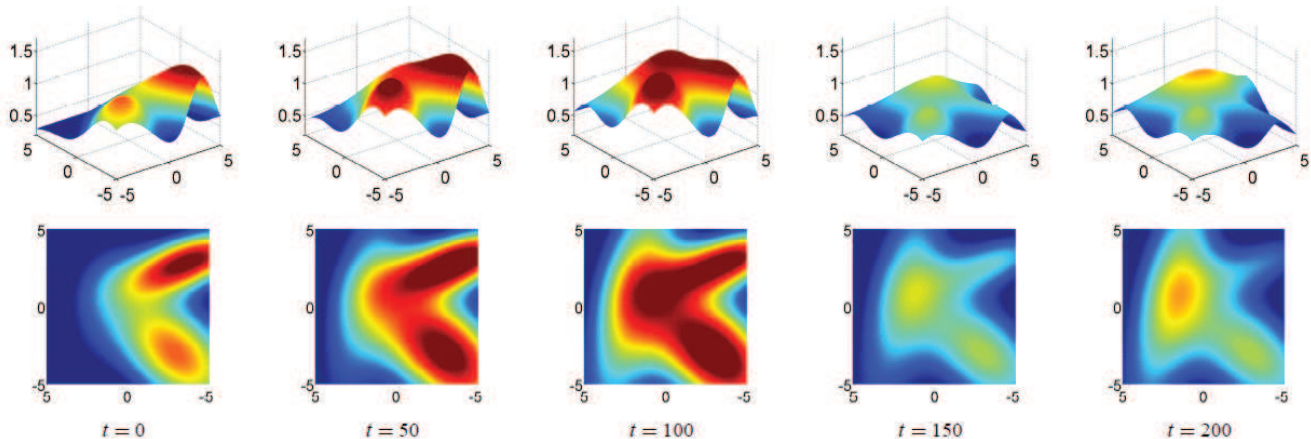}
\end{center}
\caption{\label{fig:gauss}Snapshots of a synthetic scalar field at time steps $0, 50, 100, 150$, and $200$, from left to right, respectively. [top row] Three-dimensional views. [bottom row] Two-dimensional heat-map views.}
\end{figure*}

\section{Applications toward Persistent Monitoring}\label{section:application}
In this section, we demonstrate how \copq\, and our algorithm may be applied to realistic persistent monitoring problems. Because our focus here is on estimation quality, we apply the the exact algorithm on a single mobile robot. To be extensive, we include one simulation experiment performed over a synthetic spatiotemporal field and one simulation experiment using real temperature data from $14$ weather stations in the state of Massachusetts. When it comes to applying \copq\, to persistent monitoring tasks, we must first obtain $\{w_{ij}\}$ from historical data collected over all nodes, which is a {\em learning} problem. Recall that $N_i := \{v_j \mid (v_i, v_j) \in E\}$ is the neighbor set of $v_i \in V$. We apply a linear regression model over $\psi$, {\em i.e.}, 
\begin{equation}\label{equation:psi_regression}
\psi(v_i, t) = \alpha_{0i} + \sum_{v_j \in N_i} \alpha_{ji}\psi(v_j, t), 
\end{equation}
in which $\alpha_{0i}$ and $\alpha_{ji}$'s are coefficients to be computed from historical data with $T$ sets of node value data.  \eqref{equation:psi_regression} corresponds to models such as the Gaussian Process (GP). To map these coefficients to \copq,  for each $w_{ji}$, we compute two sets of such coefficients. The first set of coefficients $\alpha_{ji}'$'s are computed assuming all of $N_i$ are visited; the second set, $\alpha_{ji}''$'s, are computed assuming $v_j$ is $v_i$'s only visited neighbor. We then compute the weight $w_{ji}$ via
\begin{equation}\label{equation:weights}
w_{ji} = \frac{\alpha_{ji}' + \alpha_{ji}''}{\sum_{k \in N_i}(\alpha_{ki}' + \alpha_{ki}'')}.
\end{equation}
Equation~\eqref{equation:weights} was chosen to balance the impact of single neighbors as well as the impact of the entire neighborhood. 

With $\{w_{ij}\}$, for a given travel budget, a utility maximizing tour can be computed. Assuming the robot collects exact values at time $t_s$ from nodes it visits, the values on nodes that are not visited by the robot are estimated as follows. Let $v_i$ be such a node and let $N_i'$ be its neighbor set such that a node $v_j \in N_i'$ has either measured or estimated value (at time $t_s$). The historical data for $v_i$ and nodes in $N_i'$ from $1 \le t \le T$ are then used to perform multiple linear regression according to 
\begin{equation}\label{equation:psi_regression_2}
\psi(v_i, t) = \beta_{0i} + \sum_{v_j \in N_i'} \beta_{ji}\psi(v_j, t).
\end{equation}
The obtained $\beta_{0i}$ and $\beta_{ji}$'s can then be used to compute the estimated $\psi'(v_i, t_s)$ using~\eqref{equation:psi_regression_2}. The process is repeated until all nodes are covered. This iterative process defines the function $\textrm{UpdateNodeEstimate}(\cdot)$ in in Algorithm~\ref{algorithm:lae}.

\subsection{Measuring a Time-Varying Scalar Field}
Our first application-driven simulation verifies the effectiveness of Equation~\eqref{equation:weights} in connecting actual scalar fields to \copq. Our experiments are performed over a synthetic scalar field generated by three two-dimensional Gaussians. These Gaussians have fixed centers but varying intensities and covariance matrices over time; we fix the centers to ensure that the spatial correlations are relatively time-invariant. The field is simulated for $201$ time steps; the snapshots of the field at time steps $0, 50, 100, 150$, and $200$ are provided in Figure~\ref{fig:gauss}. The node network used here is a $5\times 5$ randomized grids (see, {\em e.g.}, Figure~\ref{fig:5x5-ran}) scaled to the dimensions of the support of the scalar field. For each fixed travel budget, $100$ random $5\times 5$ node networks are generated. In each randomly generated network, the nodes of the network are given equal importance ({\em i.e.}, unit utility). To estimate $\alpha_{ij}'$'s and $\alpha_{ij}''$'s for computing the weights, data from the first fifty time steps were used ($T = 50$).  For running the model to obtain a utility maximizing tour, the second diagonal node from the top-left corner was used as the base node. The resulting tour is then used to obtain $\psi'(v_i, t_s)$ for $t_s = 100, 150$, and $200$, according to~\eqref{equation:psi_regression_2}. We define the {\em quality} of $\psi'(v_i, t_s)$ as 
\begin{equation}\label{equation:quality}
\frac{\sum_{t_s\in \{100, 150, 200\}}(\psi(v_i, t_s) - |\psi'(v_i, t_s) - \psi(v_i, t_s)|)}{\sum_{t_s\in \{100, 150, 200\}} \psi(v_i, t_s)}.
\end{equation}

To compare to our results, we also exhaustively search through the network for a tour starting and ending at the same base node that minimizes the same quality defined by Equation~\eqref{equation:quality} under the same travel budget. This experiment was limited to travel budgets $6$ and $8$, corresponding to tours containing up to five nodes. While our model can produce tours with many more nodes, for comparing the result, we have to exhaustively search through all tours starting from the base node to find the best one, which becomes very costly as the number of nodes is over $5$. The quality score obtained this way is denoted as ``actual quality''. The result comparing the approaches is given in Table~\ref{table:model_fidelity_synthetic}. Using the given metric, the average quality error was less than one, meaning that it was not more than the error incurred by omitting a single node. In roughly $30\%$ of the cases, the tour found using our method was identical to the one found using exhaustive tour search. 
\begin{table}[htp]
\begin{center}
	 \caption{\label{table:model_fidelity_synthetic}Model Fidelity over a synthetic scalar field}
	 \begin{tabularx}{\columnwidth}{cccc}
   \hline\hline
		Travel Budget & Model Quality & Actual quality & Relative error   \\ \hline
		6.0 & 7.16 & 7.64 & 0.48  \\ \hline
		8.0 & 8.46 & 9.38 & 0.92 \\
 	 \hline\hline
	 \end{tabularx}
\end{center}
\end{table}

\begin{figure}[htp]
\begin{center}
\begin{tabular}{ccc}
		\vspace{5mm}
    \includegraphics[width=1.0in]{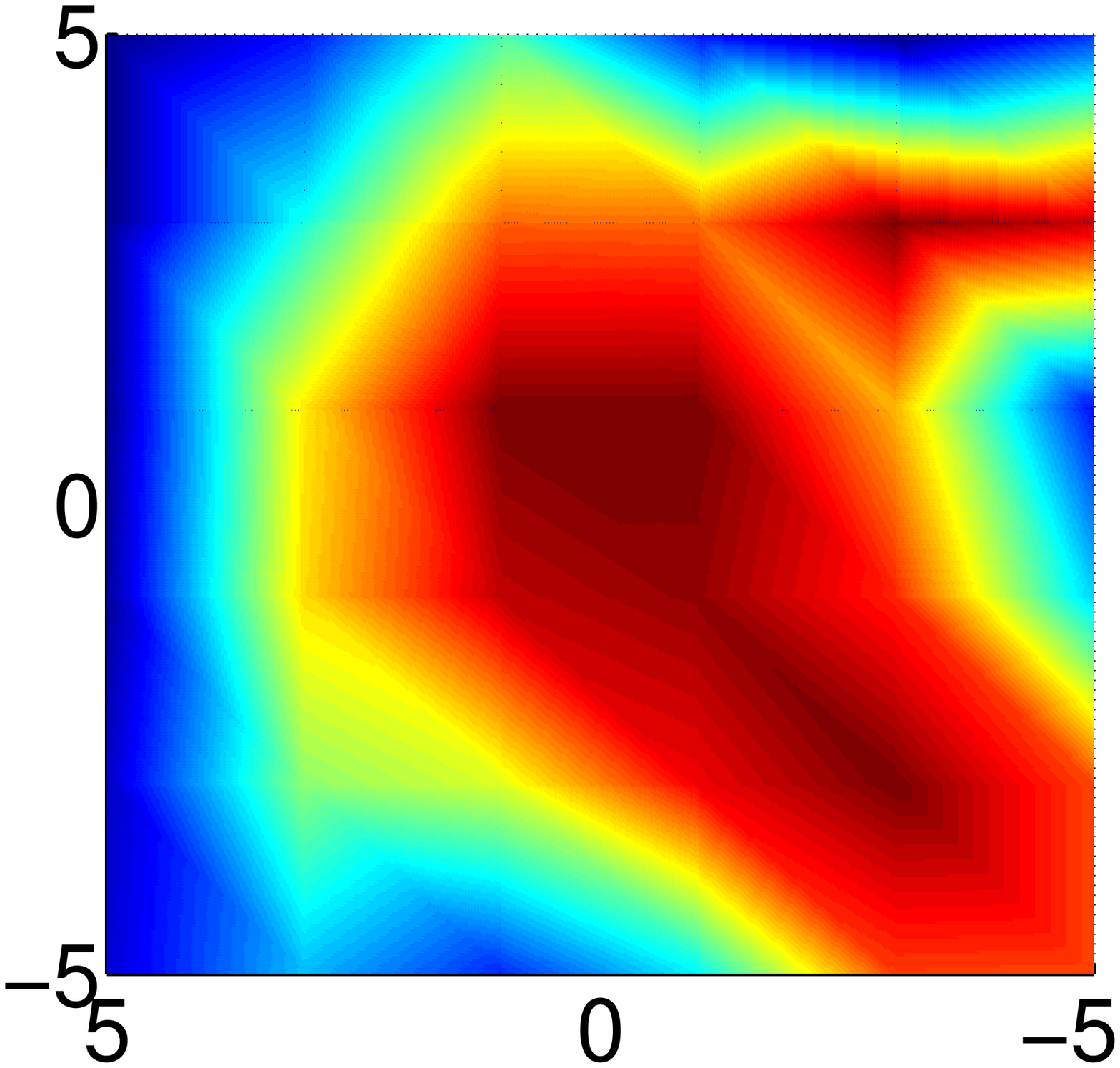} &
    \includegraphics[width=1.0in]{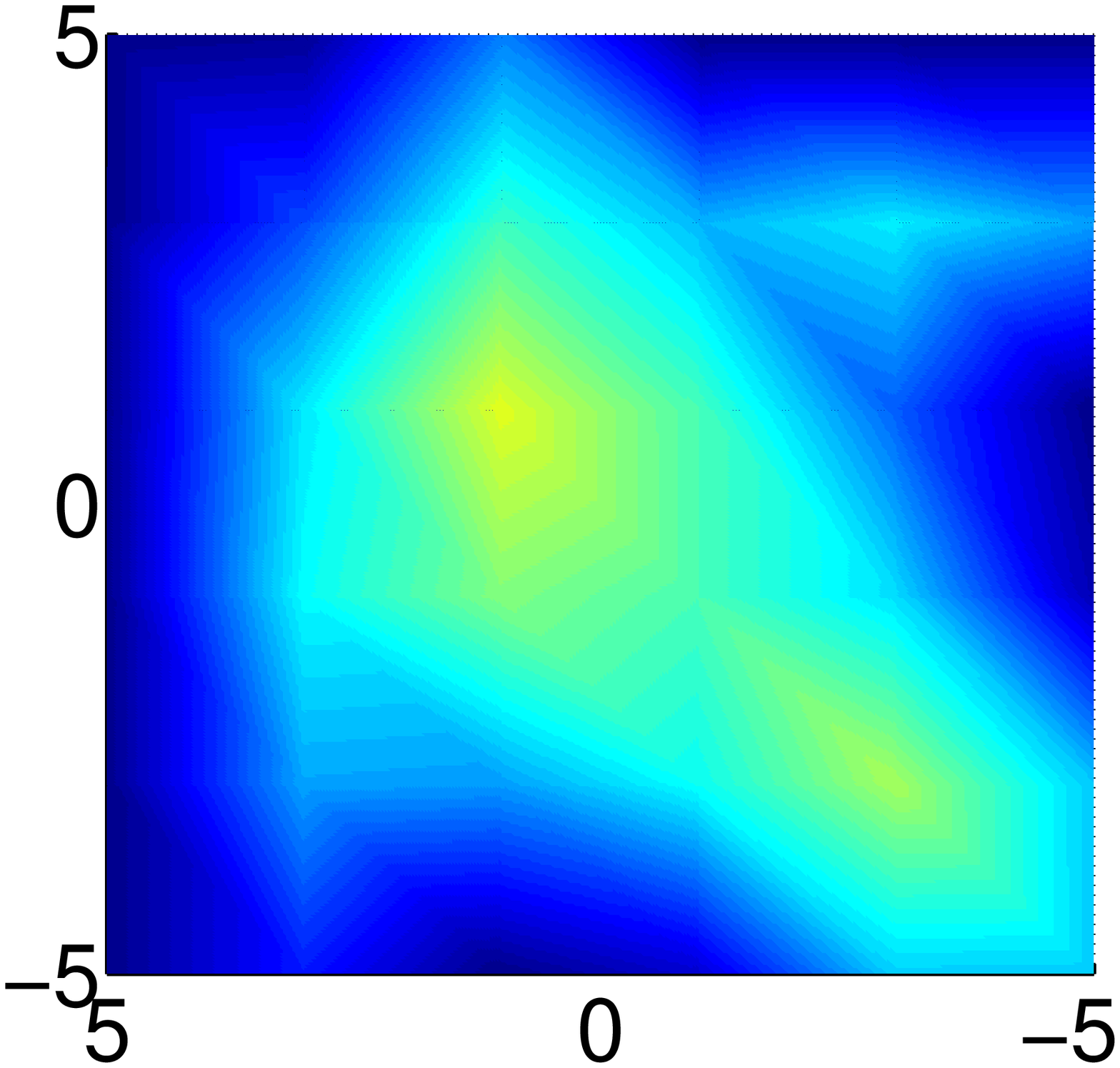} &
    \includegraphics[width=1.0in]{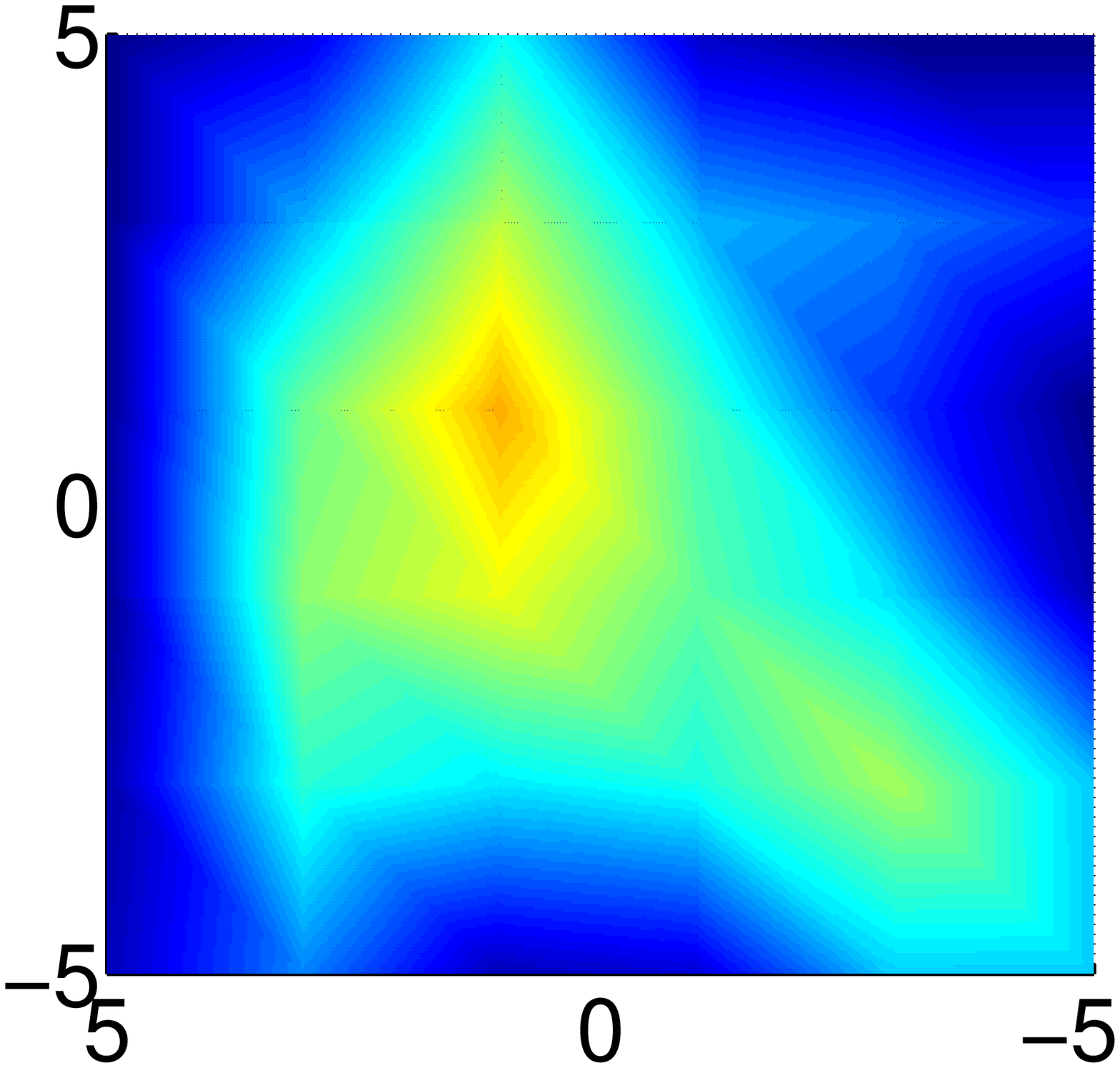} \\
		\begin{footnotesize}$t=100$\end{footnotesize}&
		\begin{footnotesize}$t=150$\end{footnotesize}&
		\begin{footnotesize}$t=200$\end{footnotesize}
\end{tabular}
\end{center}
\caption{\label{fig:simu} Estimated scalar field based on the data obtained using the MIQP model.}
\end{figure}

As a secondary and more intuitive measure of the quality of our method, we put a regular $6 \times 6$ node network fitted over the same field (Figure~\ref{fig:gauss}) and run the MIQP model such that we just have enough budget to obtain a full utility of $36$. We let the start node be the second node from the left on the first row. From the output we can then estimate values for all nodes that are not visited on the tour. We plot the much sparser survey data over the same space for time steps $100, 150$, and $200$ as shown in Figure~\ref{fig:simu}. Comparing these figures with the corresponding ones from Figure~\ref{fig:gauss}, we observe that our models provide reasonable estimation of the entire synthetic scalar field without the need to visit all the nodes. 

\subsection{Temperature Scalar Field}
\begin{figure}[htp]
\begin{center}
    \includegraphics[width=3.2in]{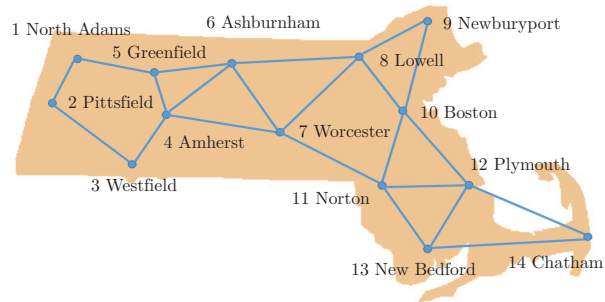}
\end{center}
\caption{\label{fig:ma} Node network of 14 weather stations in Massachusetts.}
\end{figure}

\begin{figure*}[tb!]
\begin{center}
\begin{tabular}{ccccc}
		\vspace*{3mm}
    \raisebox{-.5\height}{\includegraphics[width=1.4in]{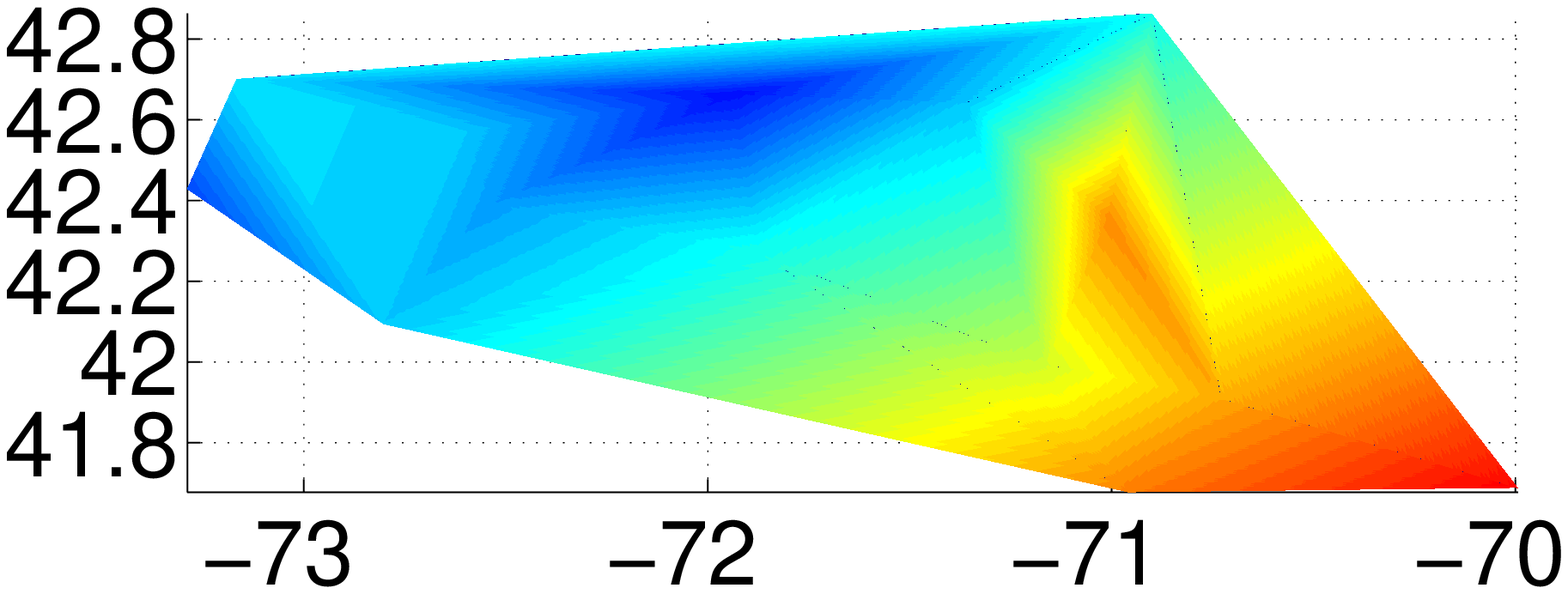}} &
    \raisebox{-.5\height}{\includegraphics[width=1.4in]{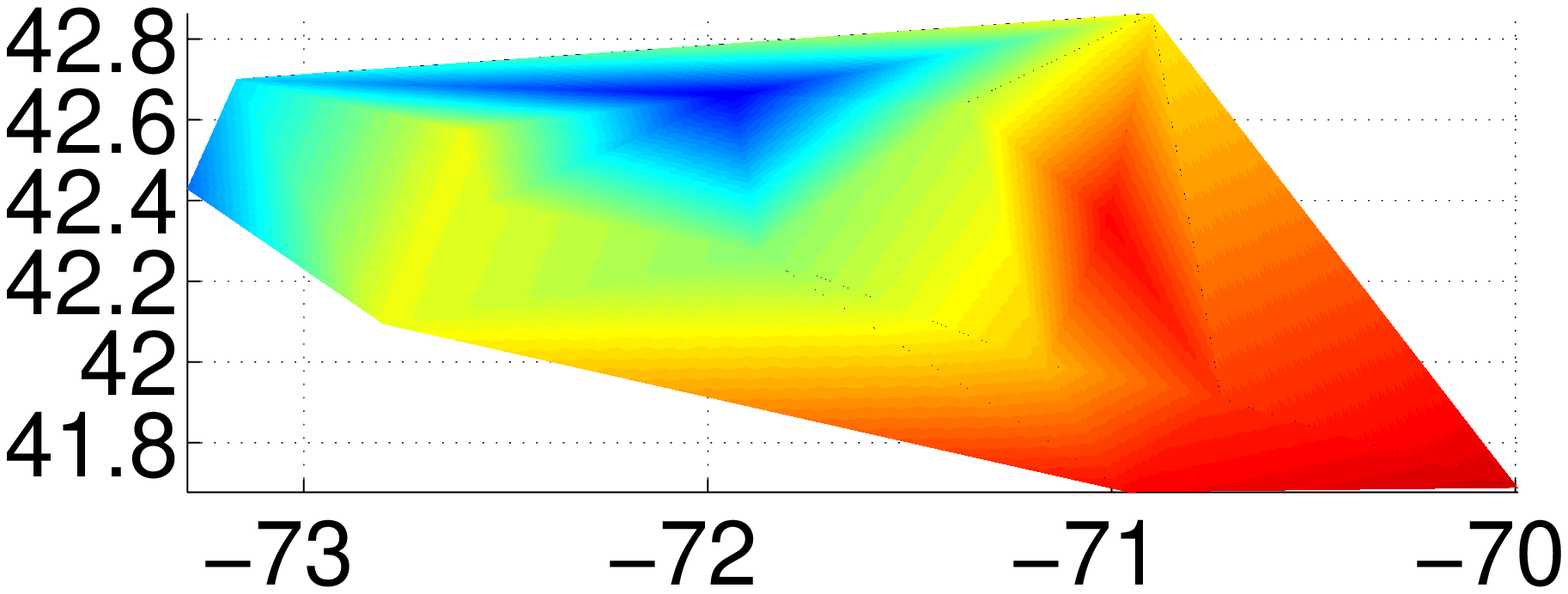}} &
    \raisebox{-.5\height}{\includegraphics[width=1.4in]{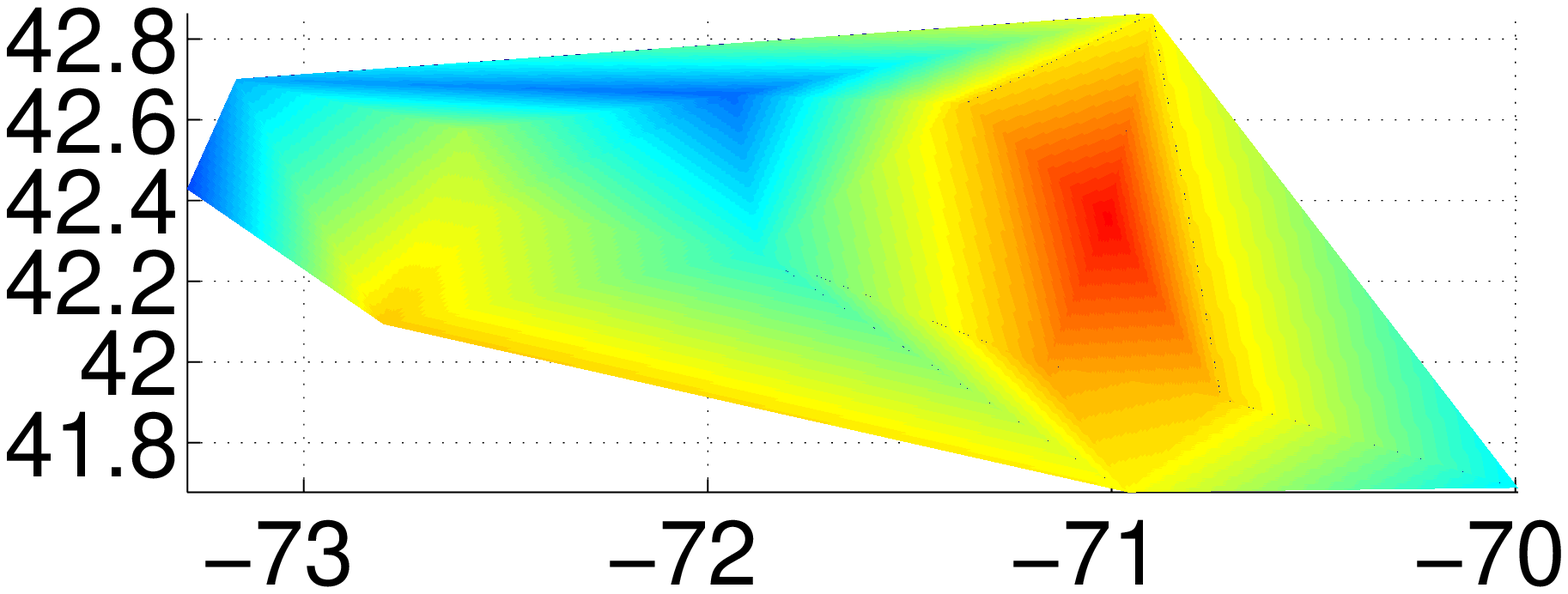}} &
    \raisebox{-.5\height}{\includegraphics[width=1.4in]{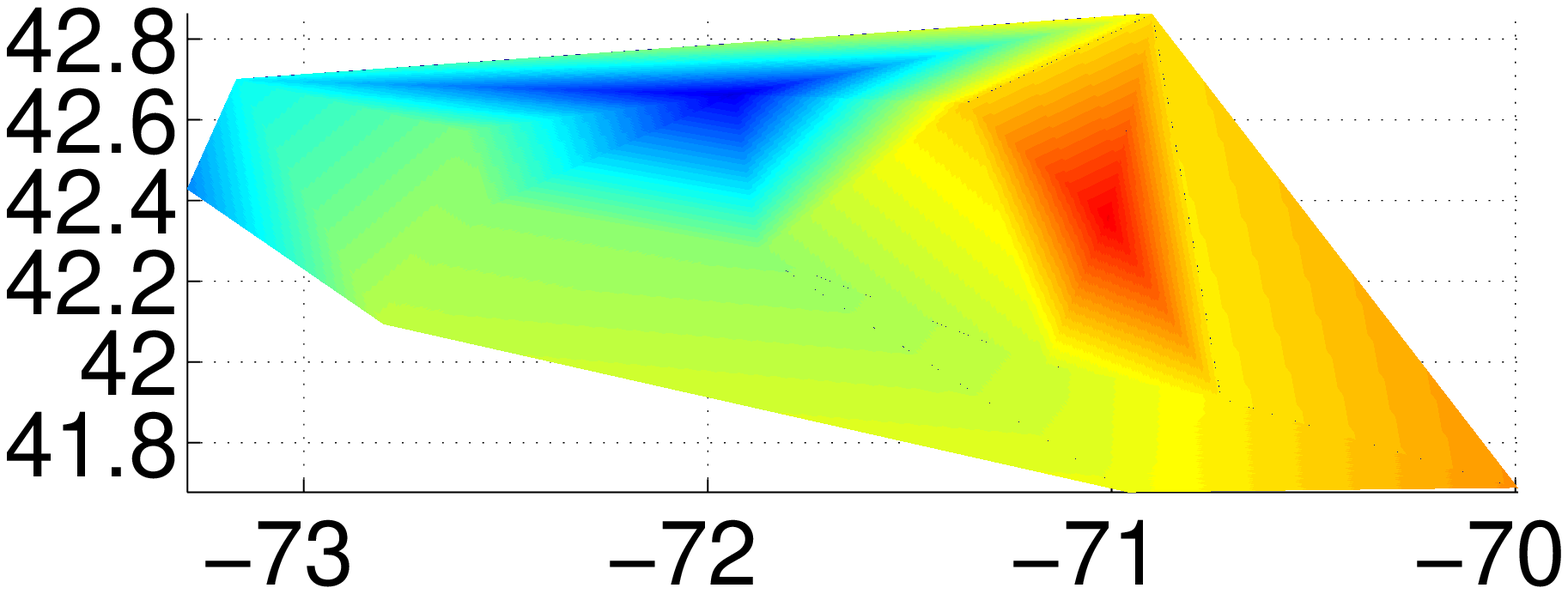}} & ground truth\\
		\vspace*{3mm}
    \raisebox{-.5\height}{\includegraphics[width=1.4in]{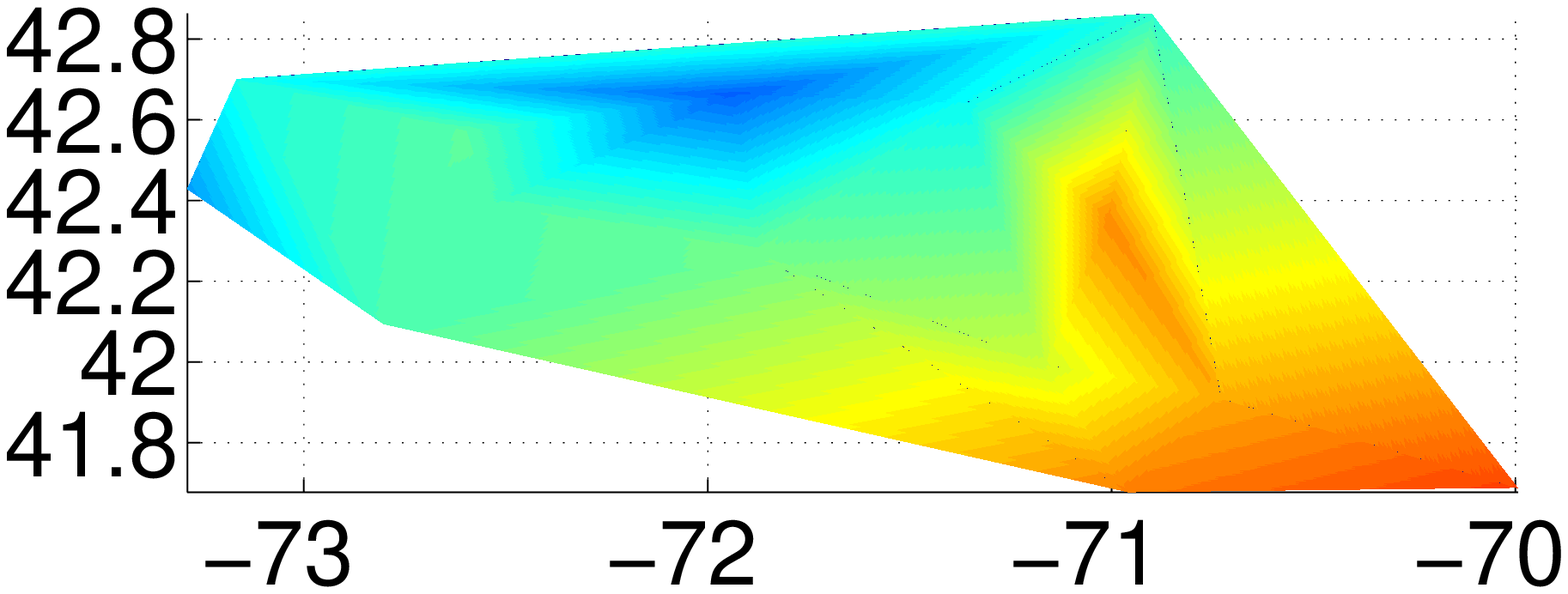}} &
    \raisebox{-.5\height}{\includegraphics[width=1.4in]{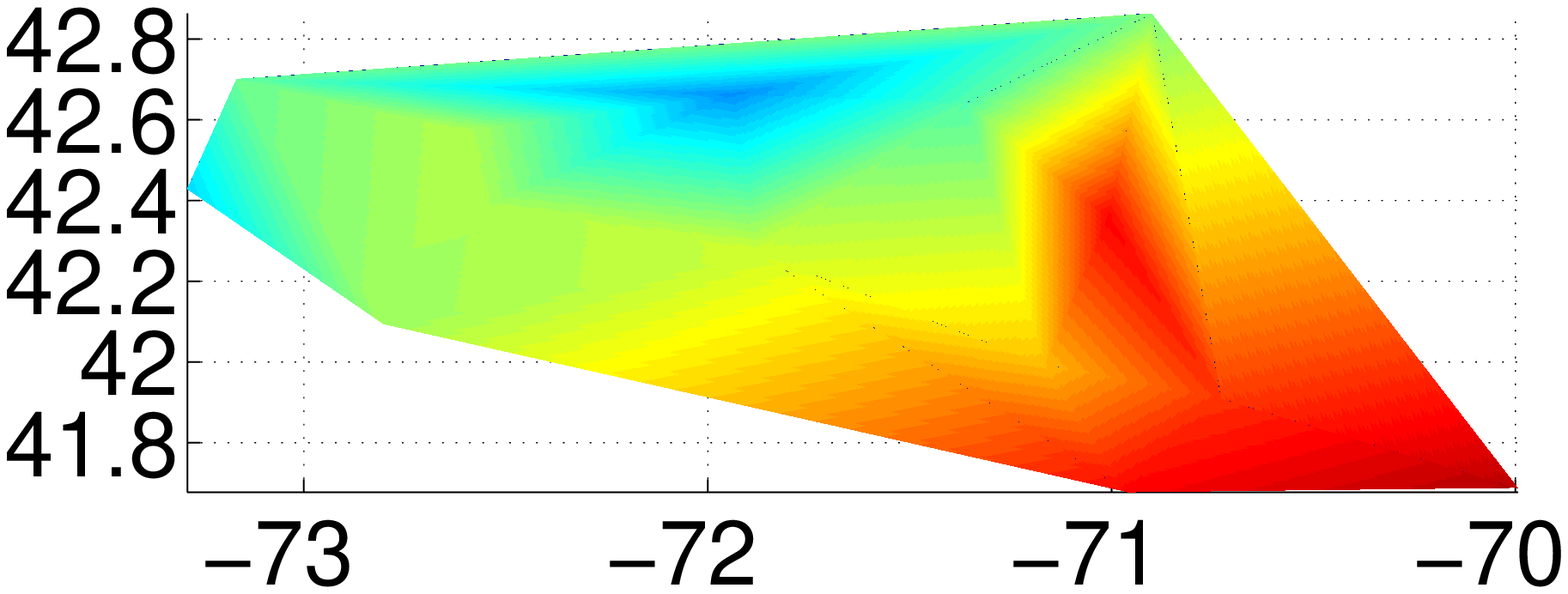}} &
    \raisebox{-.5\height}{\includegraphics[width=1.4in]{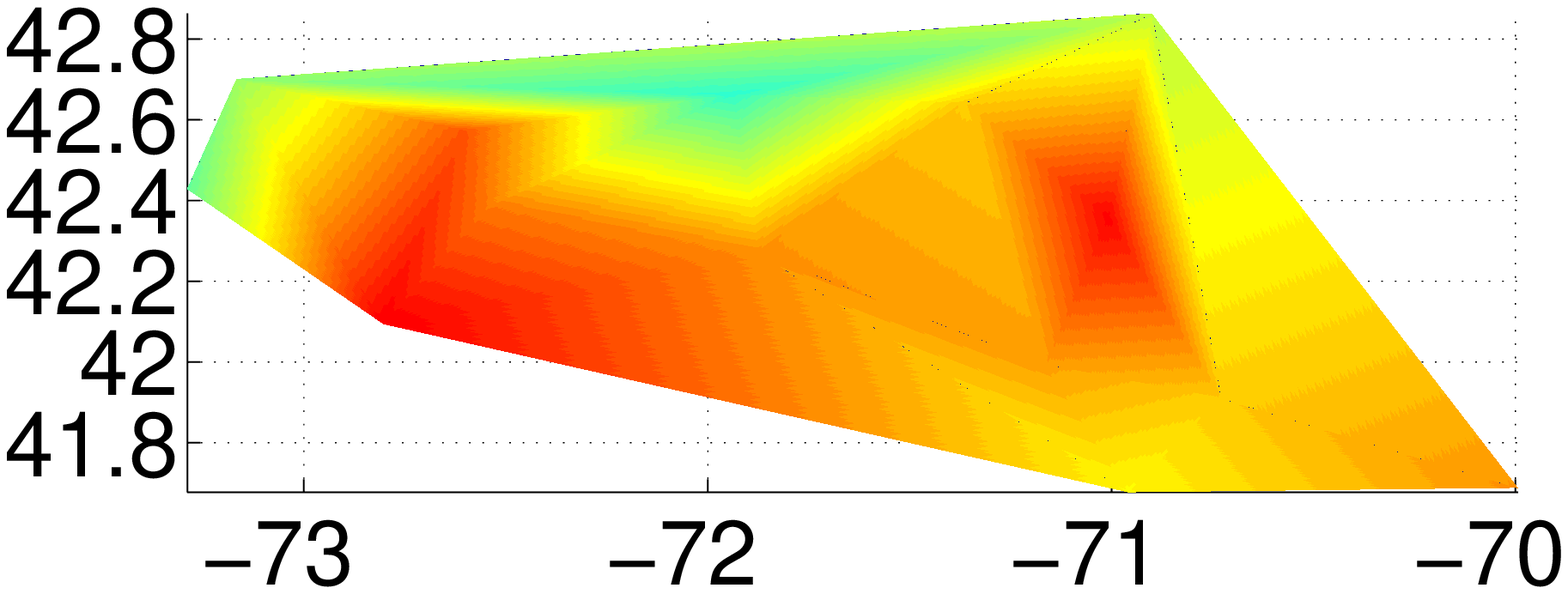}} &
    \raisebox{-.5\height}{\includegraphics[width=1.4in]{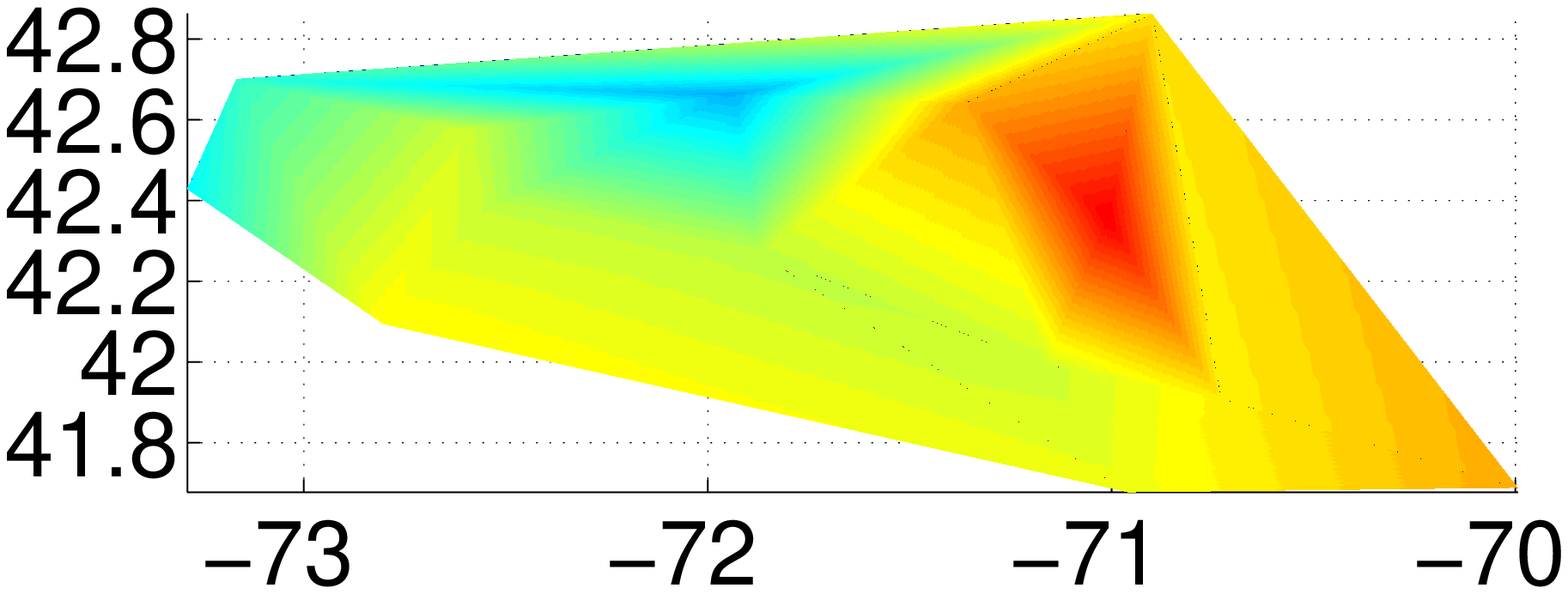}} & budget = 2.0 \\
		\vspace*{3mm}
    \raisebox{-.5\height}{\includegraphics[width=1.4in]{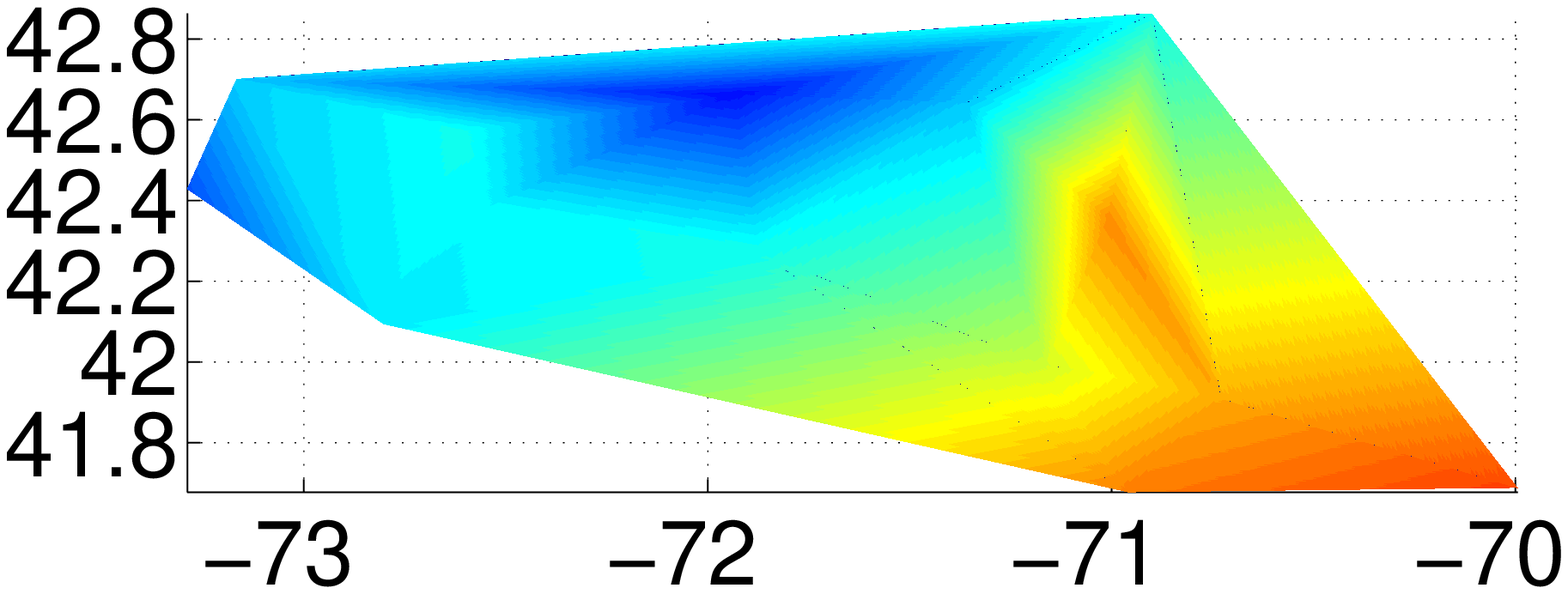}} &
    \raisebox{-.5\height}{\includegraphics[width=1.4in]{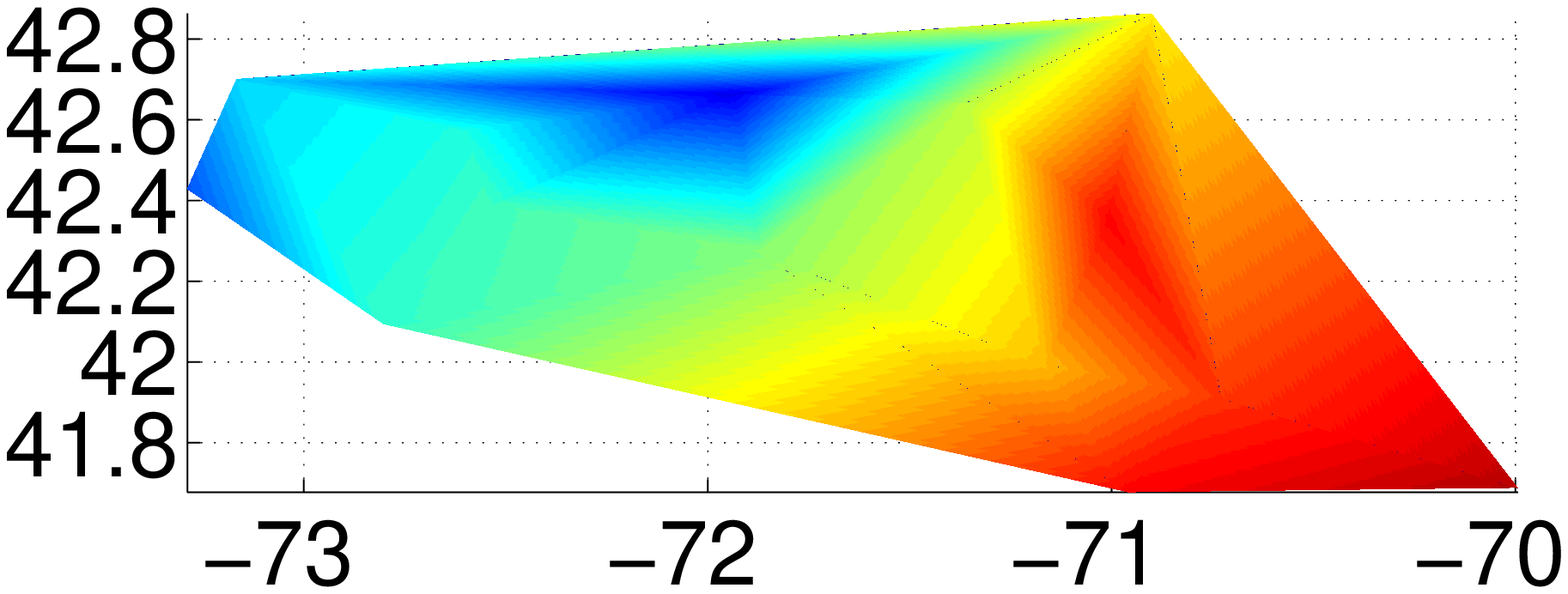}} &
    \raisebox{-.5\height}{\includegraphics[width=1.4in]{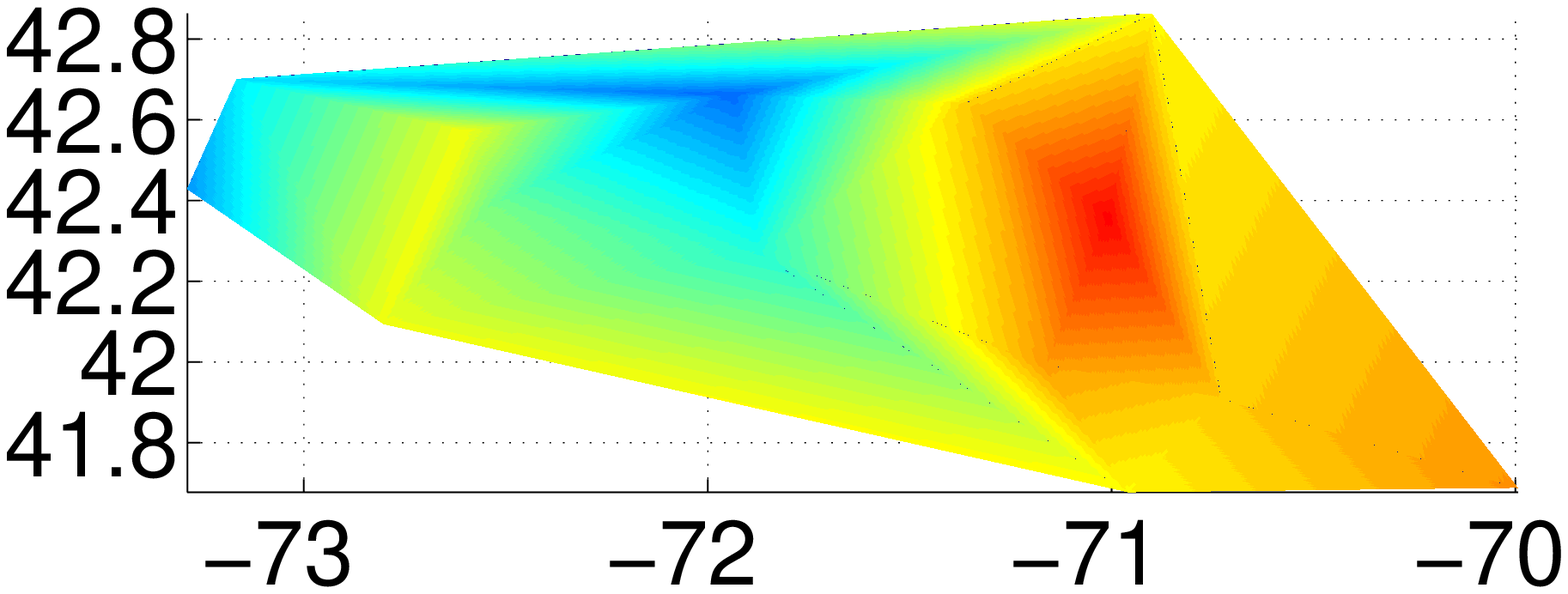}} &
    \raisebox{-.5\height}{\includegraphics[width=1.4in]{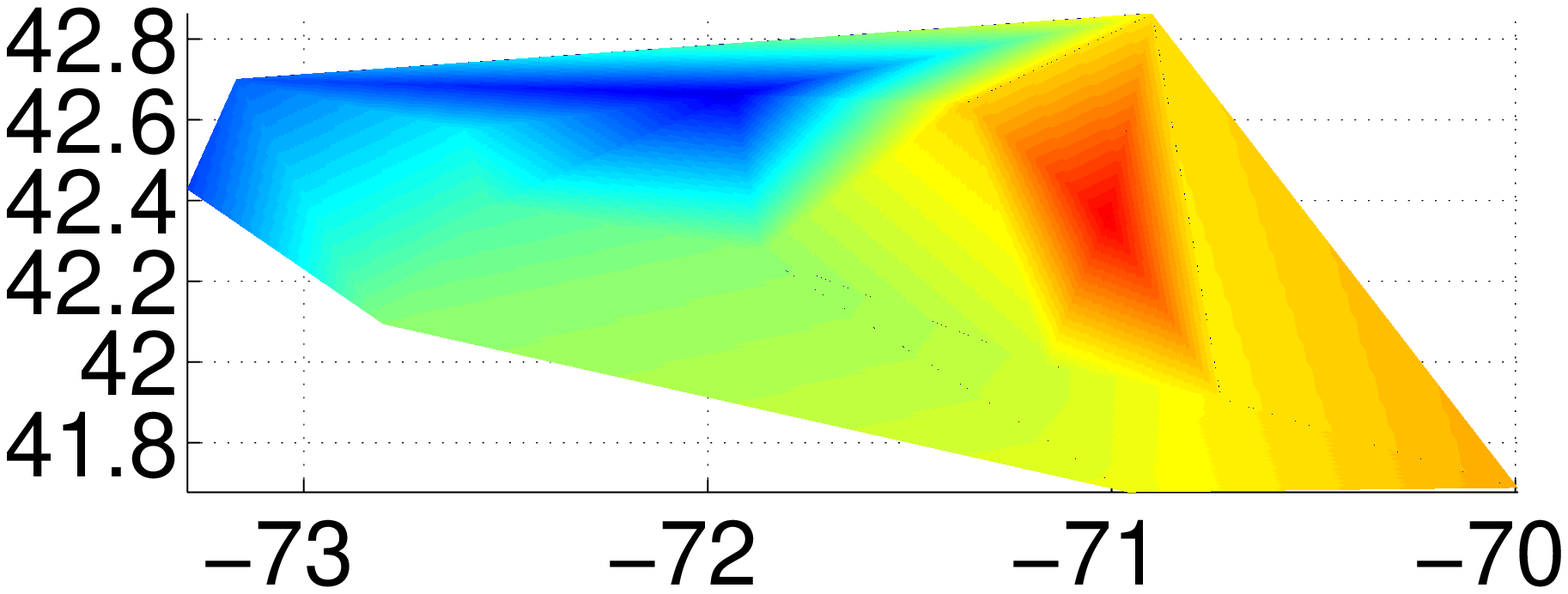}} & budget = 4.0 \\
    \raisebox{-.5\height}{\includegraphics[width=1.4in]{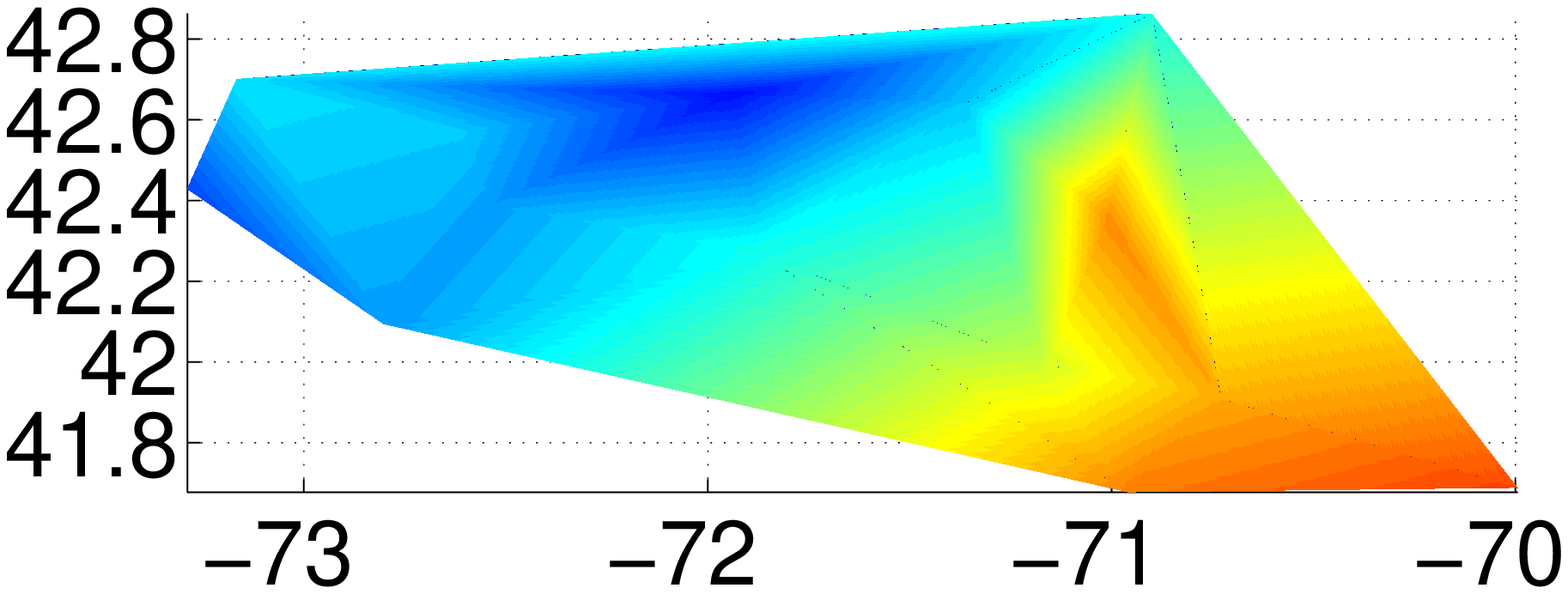}} &
    \raisebox{-.5\height}{\includegraphics[width=1.4in]{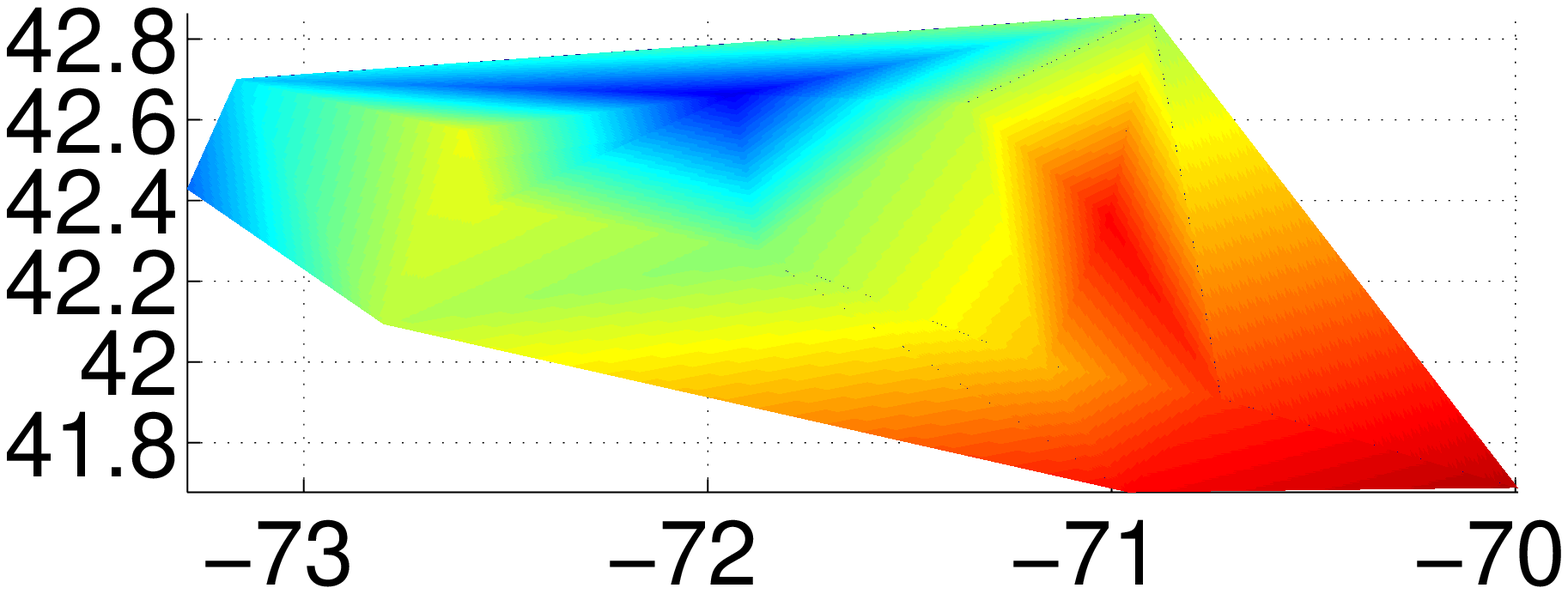}} &
    \raisebox{-.5\height}{\includegraphics[width=1.4in]{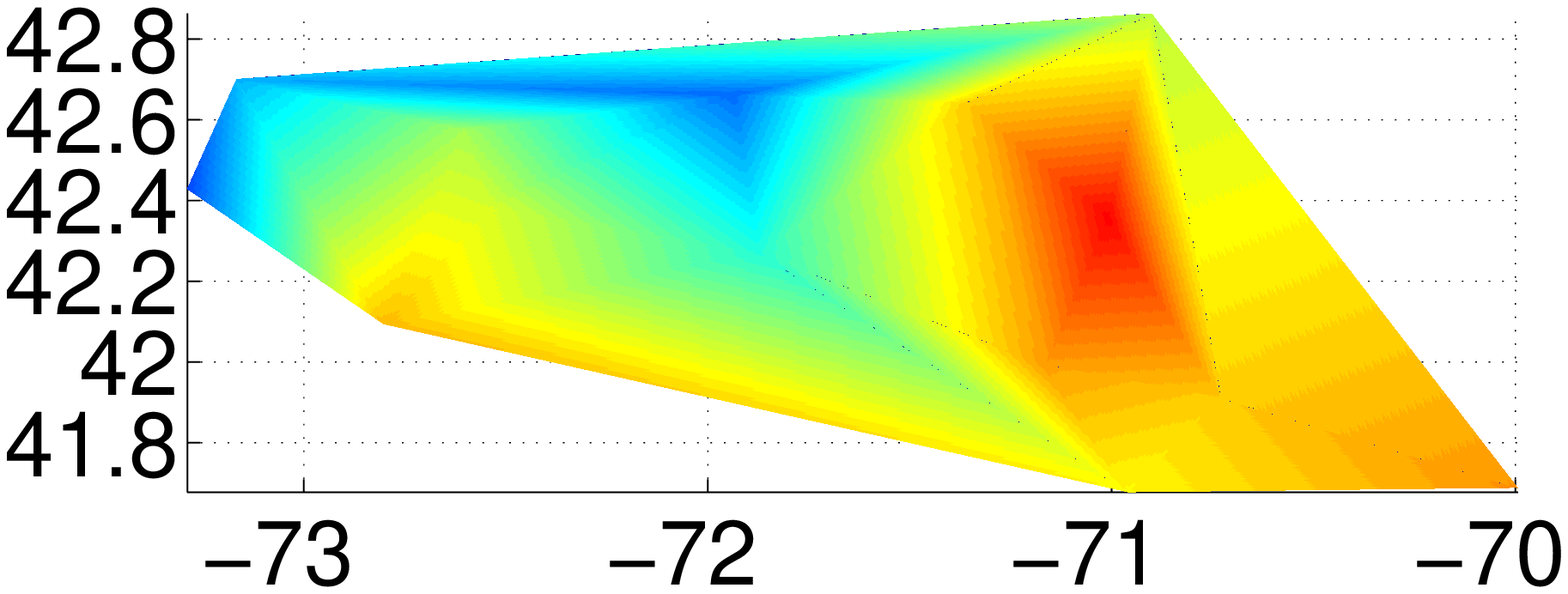}} &
    \raisebox{-.5\height}{\includegraphics[width=1.4in]{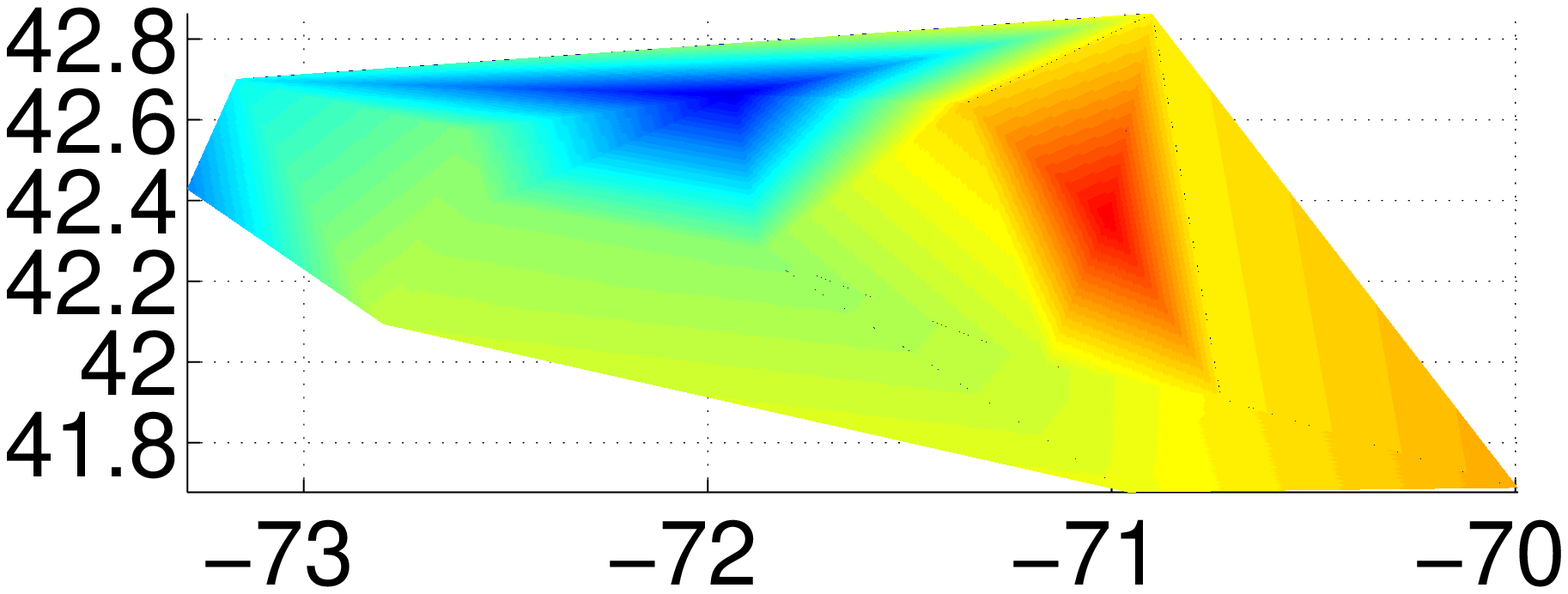}} & budget = 6.0 \par
\end{tabular}
\end{center}
\caption{\label{fig:temp} Heat maps of Massachusetts in four different seasons of a year. Note that the color-scaling for each figure is different horizontally but the color-scaling used for each column of figures is the same. [top row] Interpolations using temperature data at $14$ weather stations. [rows 2-4] Interpolations using estimated data with various budgets.}
\end{figure*}

Our second simulation works with real temperature data retrieved from National Oceanographic Data Center\footnote{\texttt{http://www.nodc.noaa.gov/General/temperature.html}}. The data consists of monthly average temperatures taken at 14 locations in the state of Massachusetts (see Figure~\ref{fig:ma}) over a 24 month period. Using methodology similar to the synthetic scalar field example, we use the first year's data as training data ({\em i.e.}, $T = 12$) and then run our algorithm to sample and estimate temperature for the next year for every three months (a total of $12/3 = 4$ sets of temperatures). Node $10$ (Boston) is selected as the base. The ground truth for these four sets is plotted in the top row of Figure~\ref{fig:temp}.

For constructing the budget, since the area of Massachusetts is relatively small, we treat it as flat and use longitudes and latitudes to measure the distance between the nodes. We run our algorithm using varying budgets from $2.0$ to $6.0$. The results are plotted using heat map in Figure~\ref{fig:temp}. Although the figures are visualizations of discrete data through interpolation, via similarity, we observe that additional budget allows better sampling and estimation quality. Quantitatively, since temperature itself is a good metric, we measure the quality of the estimation by summing up the absolute difference between actual and estimated values at each node. Then, the total error is averaged over the number of nodes. The outcome is listed in Table~\ref{table:model_fidelity_real}. At a budget of $3.0$, which is enough to visit half of the nodes, the estimation quality is already fairly good at an average error of $0.27^{\circ}$C. The error goes to less than $0.1^{\circ}$C with budget of $6.0$. On the other hand, visiting all nodes requires a budget of roughly $8.5$. 

\begin{table}[htp]
\begin{center}
	 \caption{\label{table:model_fidelity_real}Temperature Estimation Error with Respect to Budget}
	 \begin{tabularx}{\columnwidth}{cC{8mm}C{8mm}C{8mm}C{8mm}C{8mm}}
   \hline\hline
		Travel Budget & 2.0 & 3.0 & 4.0 & 5.0 & 6.0  \\ \hline
		Average Error (${}^\circ$C) & 0.46 & 0.27 & 0.22 & 0.15 & 0.08 \\
 	 \hline\hline
	 \end{tabularx}
\end{center}
\end{table}

\section{Conclusion and Future Work}\label{section:conclusion}

We introduced \cop\, (and \copq) as an extension to \op\, to address the problem of planning tours for surveying a spatially correlated field that also changes over time. Our computational experiments showed that the MIQP-based anytime algorithms for \copq\, are effective in capturing the spatial correlation among nearby nodes, indicating that \copq\, and the associated MIQP models are applicable to persistent monitoring tasks in which the mobile robots have limited travel range. 

For future work, we plan to improve our models to better capture real-world application scenarios. There are many promising directions; we mention two here. First, instead of looking at only immediate correlations as we did with \copq, it may be beneficial look at correlations of nodes that are further apart in the node network, {\em i.e.}, nodes that are in the 2- or 3-neighborhood. This extension is not trivial since the inclusion of additional nodes will certainly pose new computational challenges. We expect, however, the gain in estimation accuracy will degrade quickly as the neighborhood expands. Therefore, a 3-neighborhood is perhaps all that is needed. Second, for \copq, the current weight selection criterion for applying the model in practice is somewhat ad-hoc and has ample room for improvement. Through a more systematic approach, perhaps via statistical methods, we hope to derive more principled and systematic procedures for selecting the weights for the MIQP models to further improve its applicability. 

Furthermore, this paper only begins to address the problem of using correlations in informative path and policy planning in a discrete fashion. The dual problem to this estimation problem is a learning problem: how can we learn the correlations among the nodes so as to apply the methods from this paper? How can we carry out the learning task with limited number of mobile robots? Can we perform learning and estimation simultaneously? We will investigate these and other problems in the future.

\bibliographystyle{IEEEtranN}
\bibliography{references,jingjin}

\end{document}